\PassOptionsToPackage{dvipsnames,table}{xcolor}
\documentclass{article} \usepackage{ARXIV/iclr2027_conference,times}

\usepackage{amsmath,amsfonts,bm}

\def\eqref#1{equation~\ref{#1}}

\def\1{\bm{1}}

\DeclareMathAlphabet{\mathsfit}{\encodingdefault}{\sfdefault}{m}{sl}
\SetMathAlphabet{\mathsfit}{bold}{\encodingdefault}{\sfdefault}{bx}{n}

\usepackage{siunitx}
\usepackage{lipsum}

\usepackage{pgfplots}
\usepackage{pgfplotstable}
\pgfplotsset{compat=1.18}
\usepgfplotslibrary{groupplots}

\usepackage{tabularx}
\usepackage{adjustbox}
\usepackage{physics}
\usepackage{csquotes}
\usepackage{colortbl}
\usepackage{xcolor}
\usepackage{graphicx}
\usepackage{pifont}
\usepackage{caption}
\usepackage{xspace} 
\usepackage{hyperref}
\usepackage{url}
\usepackage{subcaption}
\usepackage{booktabs}
\usepackage{svg}
\usepackage{wrapfig}
\usepackage{placeins}
\usepackage{amssymb}
\usepackage{enumitem}

\AtEndPreamble{
    \usepackage[capitalize]{cleveref}
    \crefname{section}{Sec.}{Secs.}
    \Crefname{section}{Section}{Sections}
    \Crefname{table}{Table}{Tables}
    \crefname{table}{Tab.}{Tabs.}
}

\makeatletter
\DeclareRobustCommand\onedot{\futurelet\@let@token\@onedot}
\def\@onedot{\ifx\@let@token.\else.\null\fi\xspace}

\makeatother

\definecolor{cvprblue}{rgb}{0.21,0.49,0.74}

\hypersetup{
    colorlinks=true,
    linkcolor=cvprblue,
    citecolor=cvprblue,
    urlcolor=cvprblue,
}

\definecolor{probingbestbg}{HTML}{A8CEFF}
\definecolor{probingsecondbg}{HTML}{D9EDFF}
\newcommand{\probingscore}[2]{\cellcolor{#1}#2}
\newcommand{\probingbest}[1]{\probingscore{probingbestbg}{\textbf{#1}}}
\newcommand{\probingsecond}[1]{\probingscore{probingsecondbg}{#1}}

\newcommand{\ours}{VINO}

\newcommand{\PARbegin}[1]{\noindent {\bf #1~}}

\newcommand{\httpsurl}[1]{\href{https://#1}{\texttt{#1}}}

\definecolor{noreproduce}{gray}{0.6}

\newcolumntype{Y}{>{\centering\arraybackslash}X}
\newcolumntype{s}{>{\centering}p{2pt}}
\newcolumntype{L}{>{\raggedright\arraybackslash}X}
\newcolumntype{N}[1]{>{\raggedleft\arraybackslash}p{#1}}

\definecolor{c_red}{HTML}{ea4335}
\definecolor{c_blue}{HTML}{4285f4}
\definecolor{c_orange}{HTML}{ff6d01}
\definecolor{c_green}{HTML}{34a853}
\definecolor{c_purple}{HTML}{9900ff}
\definecolor{c_yellow}{RGB}{241,191,66}
\definecolor{c_other}{HTML}{000000}

\definecolor{defaultgray}{gray}{0.78}

\newcommand{\defaultrow}{\rowcolor{defaultgray}}

\definecolor{dino}{HTML}{DBEAFE}
\definecolor{dino_border}{HTML}{51A2FF}
\definecolor{backbone}{HTML}{FEF3C6}
\definecolor{backbone_border}{HTML}{FFB900}
\makeatletter
\newcommand*\fsize{\dimexpr\f@size pt\relax}\makeatother

\usepackage[normalem]{ulem}
\usepackage{multirow}

\makeatletter
\newcommand\notsotiny{\@setfontsize\notsotiny{6.5}{7.5}}
\newcommand\notscriptsize{\@setfontsize\notscriptsize{7.5}{8.5}}
\newcommand\notfootnotesize{\@setfontsize\notfootnotesize{8}{9}}
\makeatother

\title{\centering\normalfont\ours{}: Revisiting 3D Representation Learning with Volume Transformers}

\author{\makebox[\dimexpr\textwidth-2\tabcolsep\relax][s]{Adrian Kruse\textsuperscript{1*} \hfill
Kadir Yilmaz\textsuperscript{1*} \hfill
Daan de Geus\textsuperscript{2} \hfill
Francis Engelmann\textsuperscript{3} \hfill
Bastian Leibe\textsuperscript{1}} \\[0.5em]
\makebox[\dimexpr\textwidth-2\tabcolsep\relax][c]{\small\textsuperscript{1}RWTH Aachen University \enspace
\textsuperscript{2}Eindhoven University of Technology \enspace
\textsuperscript{3}USI Lugano \enspace
\textsuperscript{*}Equal contribution} \\[0.5em]
\makebox[\dimexpr\textwidth-2\tabcolsep\relax][c]{\normalfont\httpsurl{vision.rwth-aachen.de/VINO}}
}

\iclrfinalcopy 
\begin{document}

\vspace*{-2\baselineskip}
\maketitle
\fancyhead{}
\renewcommand{\headrulewidth}{0pt}
\vspace*{-1\baselineskip}

\begin{abstract}
Self-supervised learning has given rise to vision foundation models that learn general-purpose image representations transferable across downstream tasks, yet representation learning for 3D scenes has not achieved comparable success.
Motivated by this, we identify key limitations of current 3D representation learning approaches and revisit their core design choices.
We first replace the commonly used hierarchical PTv3 backbone with the Volume Transformer, a ViT-like architecture for 3D scenes.
We then redesign two core objectives of modern approaches built around student--teacher networks: matching representations across views and predicting masked content from visible context.
For the first objective, we separate the roles of teacher and student: the teacher provides clean, consistent targets while the student learns invariance to stronger perturbations.
For the second objective, we prevent the student from accessing the masked geometry and instead require it to infer missing representations from visible context only.
Finally, we redesign 2D-to-3D knowledge distillation, replacing direct regression of view-dependent image features with view-consistent supervision.
Together, these changes form \ours{}, a framework that learns semantic representations of 3D scenes without labels.
\ours{} substantially improves dense linear probing across indoor and outdoor benchmarks and transfers strongly to out-of-domain settings.
\end{abstract}

\section{Introduction}
Learning general-purpose representations that transfer across diverse downstream tasks is a central goal of representation learning.
Self-supervised learning (SSL) has made remarkable progress toward this goal in computer vision, giving rise to vision foundation models (VFMs)~\citep{zhou2021ibot,oquab2023dinov2,simeoni2025dinov3} that have advanced the state of the art across downstream tasks~\citep{lin2025depthanything3,kerssies2025eomt,simeoni2025dinov3}.
A key ingredient behind this progress is \emph{self-distillation}, in which a student network is trained to predict targets produced by a teacher, which itself is updated as an exponential moving average (EMA) of the student.

\begin{figure*}[t]
\centering

\newcommand{\tes}[1]{\includegraphics[width=0.23\textwidth]{ARXIV/img/teaser/#1.png}}
\newcommand{\qh}[1]{\makebox[0.19\textwidth][c]{\small\textbf{#1}}}

\setlength{\tabcolsep}{2pt}

\begin{tabular}{cccc}
\qh{\shortstack{Input Scene\\[-1pt]\normalfont\scriptsize \vphantom{(Backbone: PTv3)}}} &
\qh{\shortstack{Sonata\\[-1pt]\normalfont\scriptsize (Backbone: PTv3)}} &
\qh{\shortstack{Sonata\\[-1pt]\normalfont\scriptsize (Backbone: Volt)}} &
\qh{\shortstack{\ours{}\\[-1pt]\normalfont\scriptsize (Backbone: Volt)}} \\

\tes{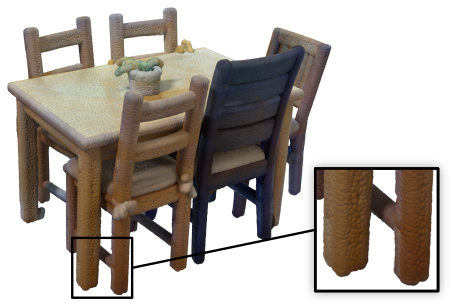} &
\tes{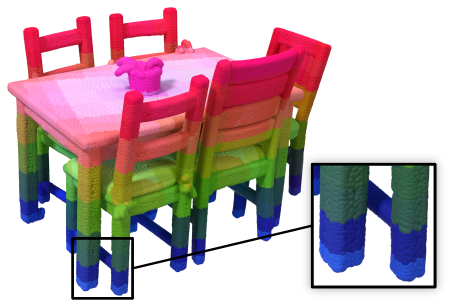} &
\tes{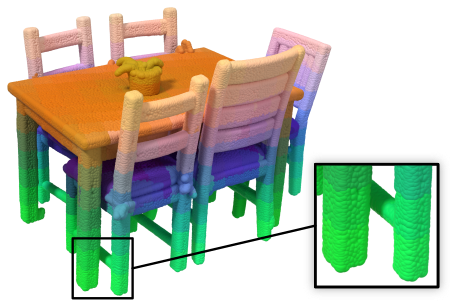} &
\tes{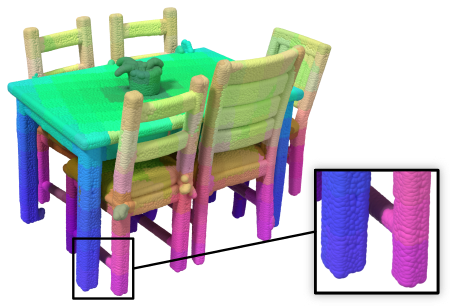} \\[2pt]

\tes{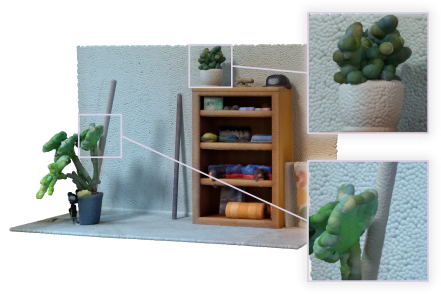} &
\tes{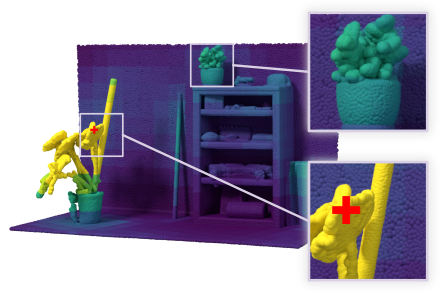} &
\tes{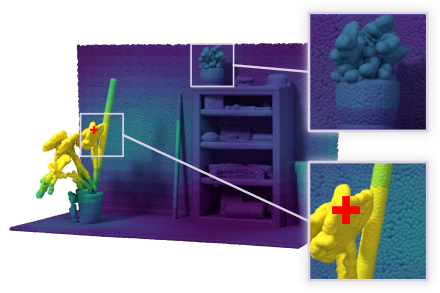} &
\tes{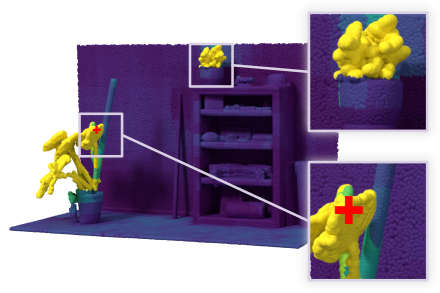}
\end{tabular}

\vspace{-5pt}
\caption{\textbf{Self-supervised 3D Representations.}
We propose \ours{}, a framework for learning semantic representations of 3D scenes with no labels. Trained on 3D data only, \ours{} captures semantic similarity better than Sonata~\citep{wu2025sonata}, as shown in PCA projections \emph{(top)} and cosine-similarity maps \emph{(bottom)}. Sonata yields qualitatively similar representations across backbones, whereas \ours{} better distinguishes semantically different but geometrically similar regions, for example, chair and table legs, and captures semantic similarity across geometrically different instances, as illustrated by one plant strongly activating the other.}
\label{fig:teaser}
\vspace{-10pt}
\end{figure*}

Recent SSL methods for 3D scenes~\citep{wu2025sonata,abdelsamad2026dos,yang2026pointins} adapt this self-distillation paradigm to 3D point clouds, but the resulting representations do not yet reach the quality observed for 2D images.
In 2D, frozen SSL encoders equipped with a linear prediction head can approach the performance of fully supervised models of comparable size~\citep{oquab2023dinov2,simeoni2025dinov3}.
In contrast, recent 3D SSL methods still substantially underperform supervised models, despite training on significantly more data.
Motivated by this gap, we identify and address limitations of current 3D self-distillation in both the backbone architecture and the training objectives, yielding semantically richer 3D representations (\cref{fig:teaser}).
We further augment the pipeline with a redesigned 2D-to-3D knowledge distillation objective.
Together, these components form \emph{\ours{}}, a novel framework for \emph{volumetric distillation with no labels}.

We first reconsider the 3D backbone architecture.
Modern self-distillation methods are developed around Vision Transformers (ViTs)~\citep{dosovitskiy2021vit}, which operate on sequences of image tokens, whereas 3D SSL methods predominantly rely on the hierarchical Point Transformer V3 (PTv3) encoder~\citep{wu2024ptv3}, which operates on multi-resolution features.
We therefore replace PTv3 with the state-of-the-art \emph{Volume Transformer} (Volt)~\citep{yilmaz2026volume}, a ViT-like architecture operating directly on volumetric tokens.
This change removes several architecture-specific mechanisms required for compatibility with PTv3 and already yields a moderate improvement.
Building on this stronger foundation, we next revisit the two core objectives of self-distillation and examine how they should be formulated for sparse 3D scenes.

\begin{figure*}[t]
    \centering

    \setlength{\tabcolsep}{2pt}

\newcommand{\viewimg}[2][0.19\textwidth]{\includegraphics[
            width=#1,
            trim=250 0 300 200,
            clip
        ]{#2}}

    \begin{tabular}{ccccc}
        \small\textbf{Sonata Global} & 
        \small\textbf{Sonata Masked} &
        \small\textbf{\ours{} Global} &
        \small\textbf{\ours{} Masked} &
        \small\textbf{\ours{} Local} \\[3pt]
\viewimg{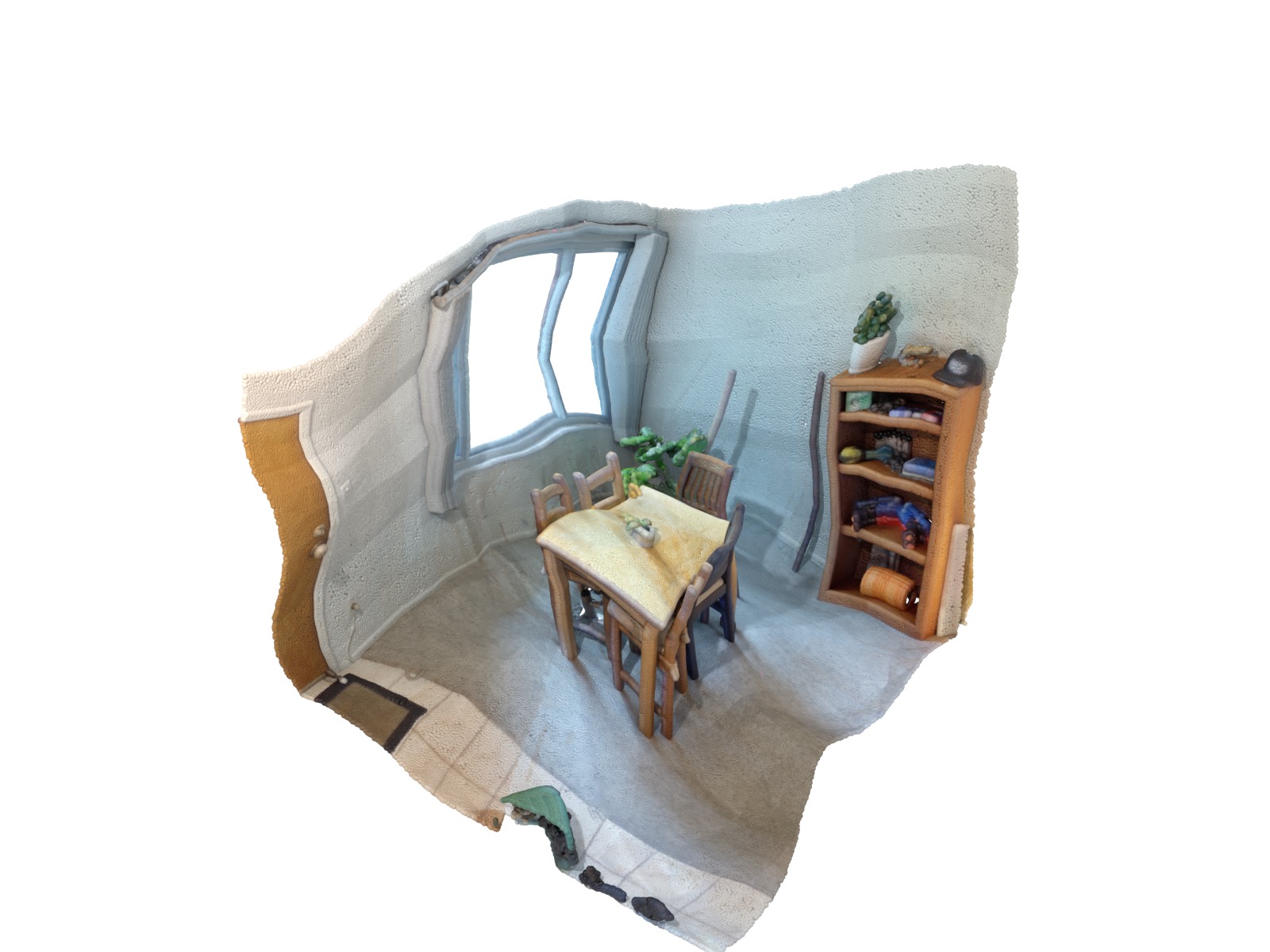} &
        \viewimg{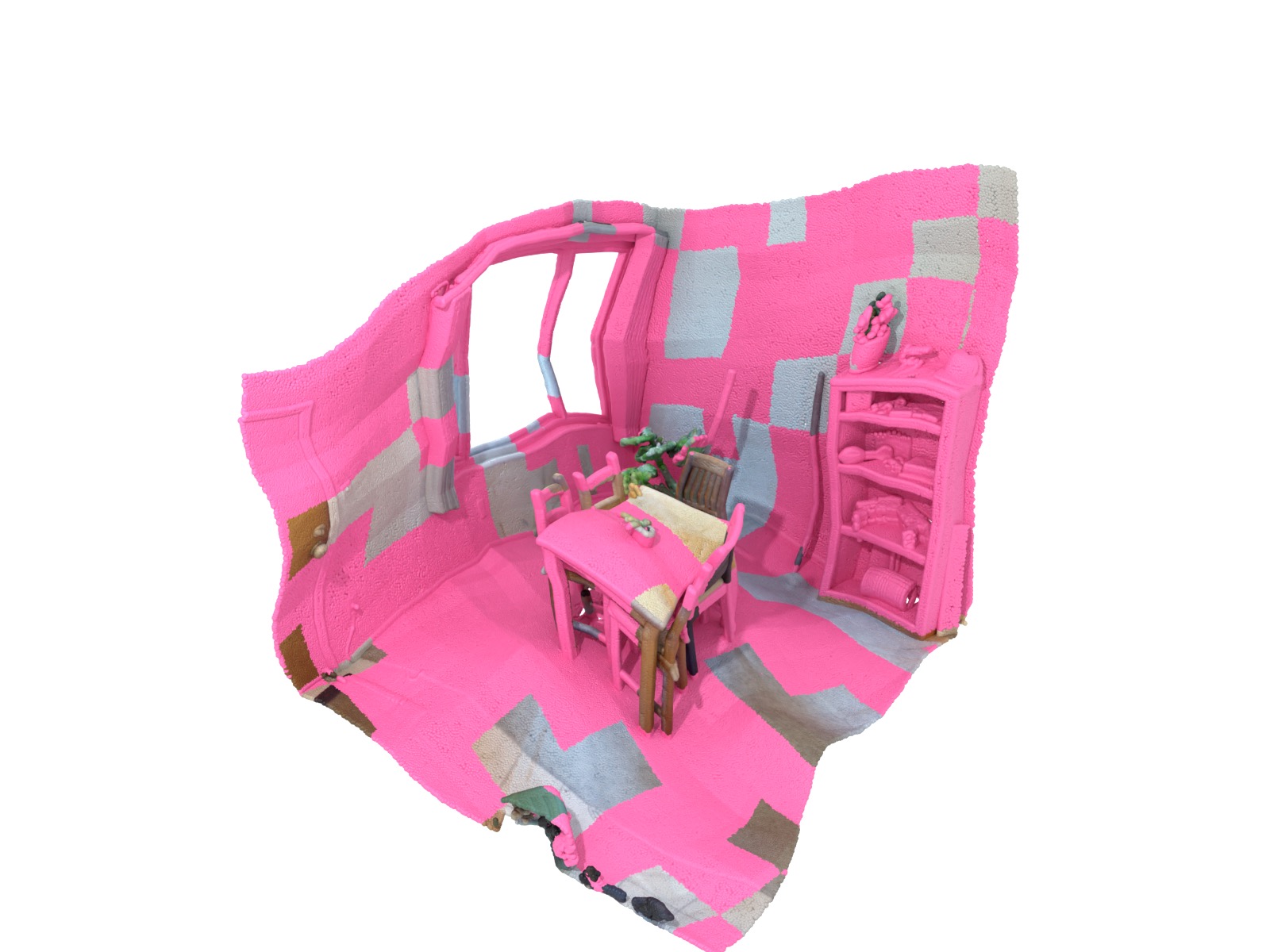} &
        \viewimg{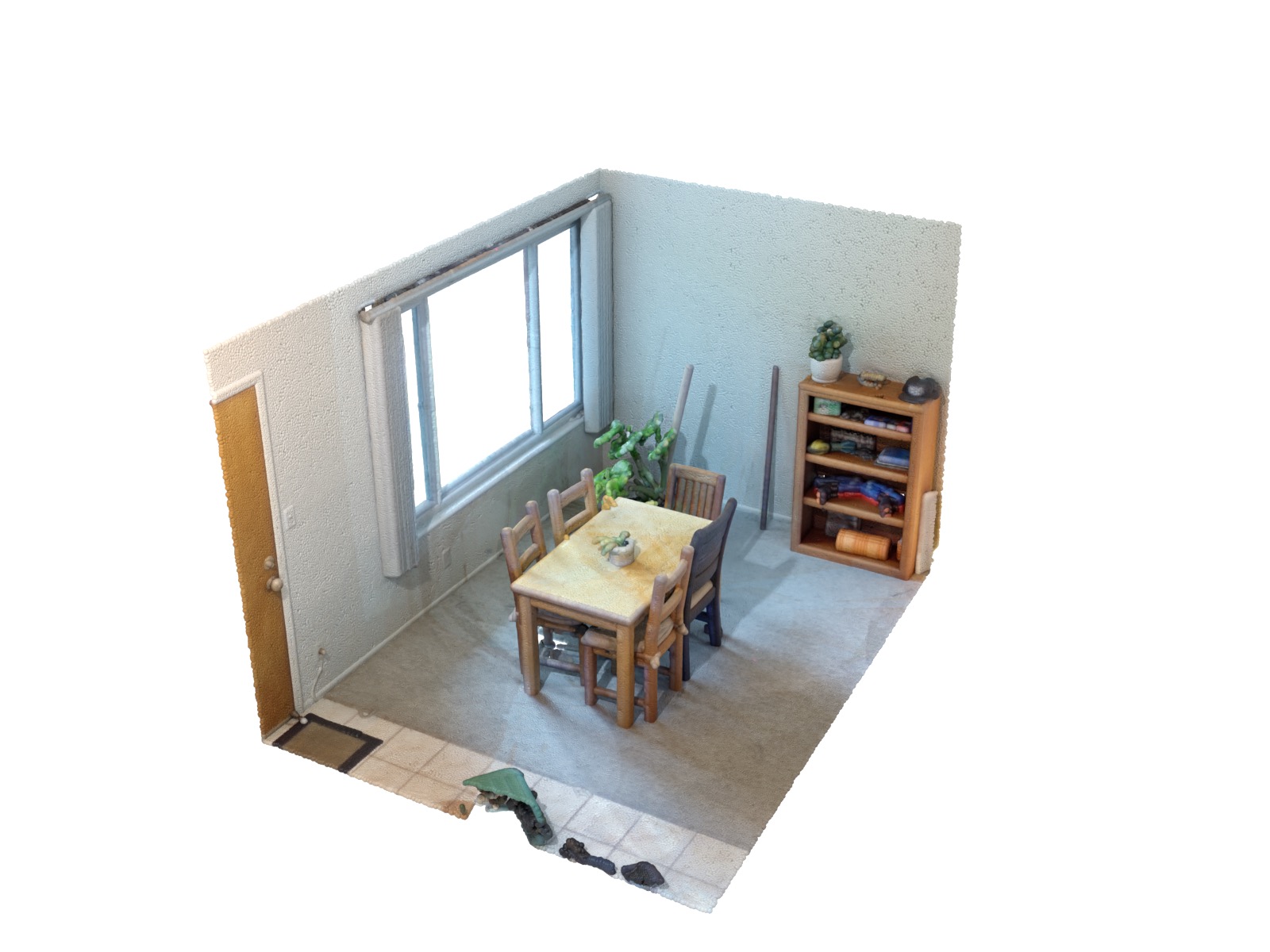} &
        \viewimg{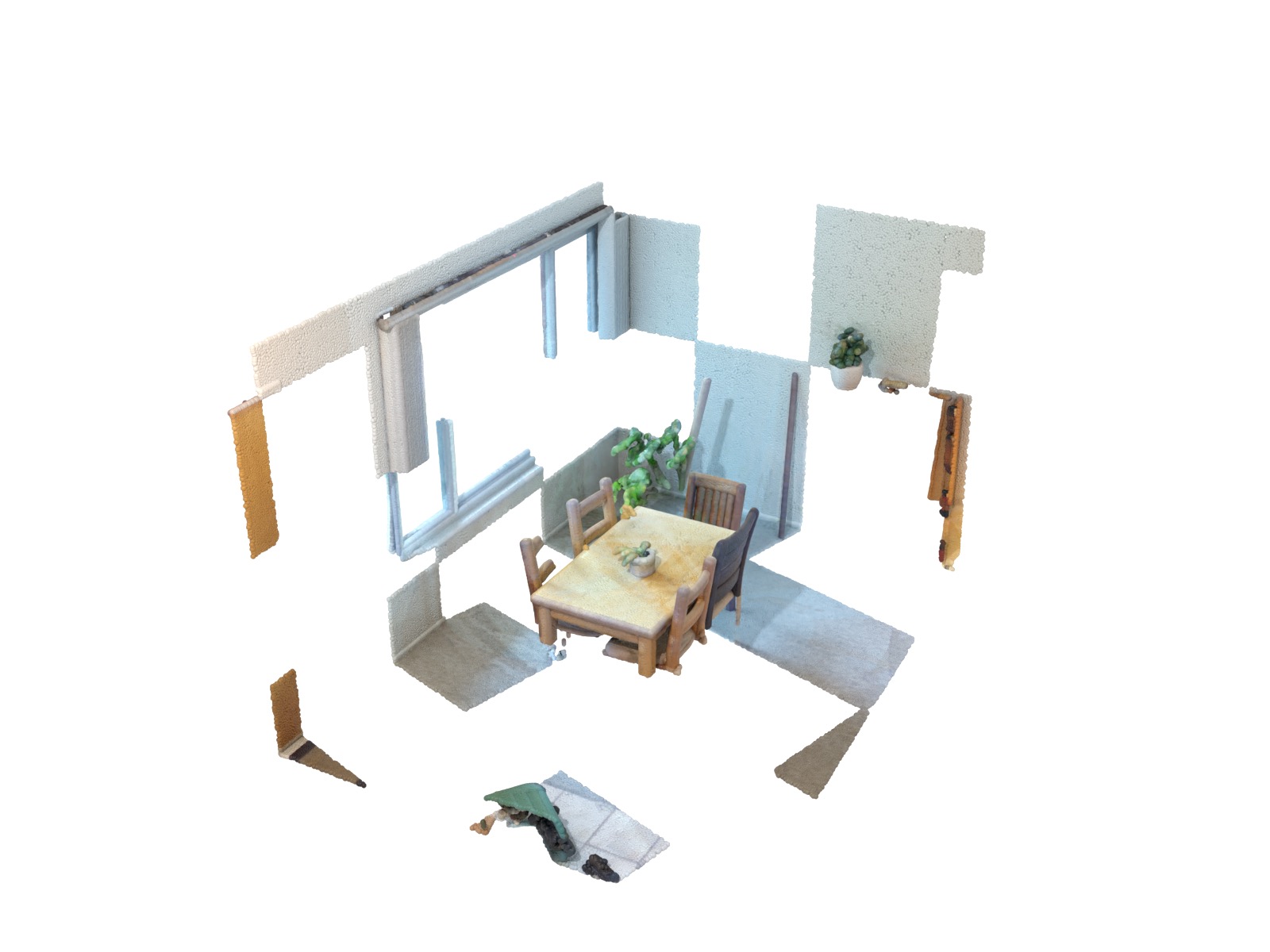} &
        \viewimg{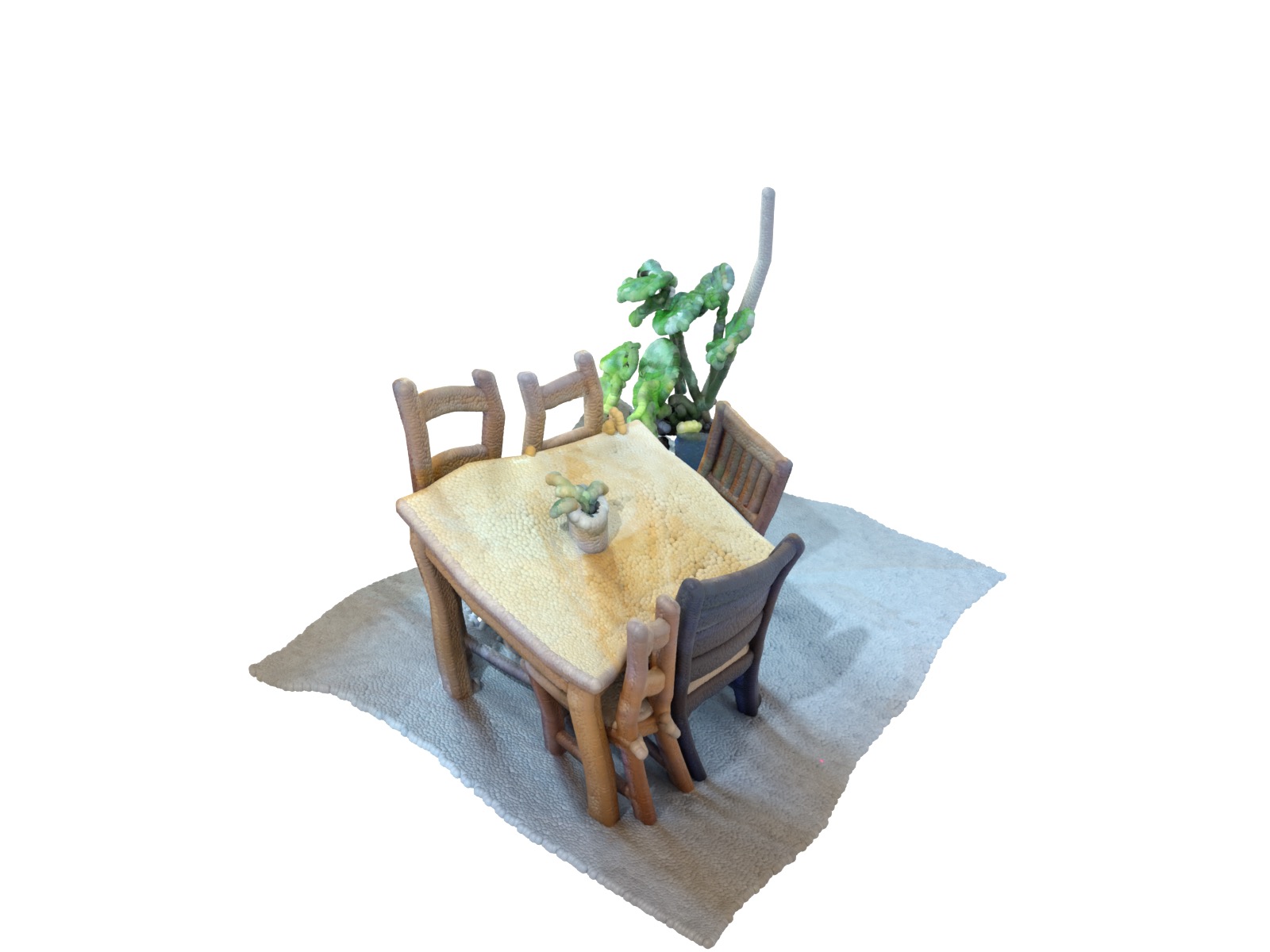}
\end{tabular}
    \vspace{-5pt}
    \caption{
    \textbf{Comparison of Training Views.}
    Sonata~\citep{wu2025sonata} geometrically corrupts all views and retains masked tokens in the student encoder. In contrast, \ours{} keeps the teacher's global views clean, removes masked tokens from the student encoder, and applies geometric corruptions only to local views.
} \label{fig:augmented_views}
    \vspace{-10pt}
\end{figure*}

The first objective trains the student encoder, through a projection head, to match the teacher's output across differently augmented views of the same scene.
Existing approaches typically apply the same set of augmentations to both the teacher and student views.
However, this exposes the teacher to the same strong perturbations as the student, potentially degrading the quality of its targets.
We instead introduce an asymmetric formulation in which the teacher receives clean global views while the student processes strongly augmented local views (see \cref{fig:augmented_views}).
This asymmetry also motivates a reconsideration of how the teacher should construct targets.
Conventionally, the teacher's projection head is obtained as an EMA copy of the student's.
Under our formulation, however, teacher and student operate on different view distributions, making this coupling undesirable.
Inspired by \cite{darcet2025capi}, we \emph{decouple target construction from student prediction} and learn the teacher's projection head independently by enforcing multi-view consistency between teacher views.

The second objective, masked modeling, trains the student encoder, typically through a separate projection head, to predict the teacher’s output at masked locations using the surrounding visible context.
Mirroring masked image modeling, existing scene-level 3D methods commonly replace masked voxel features with learned embeddings but keep the masked voxels at their original locations in the sparse grid.
In sparse 3D scenes, however, the occupancy of voxels itself encodes geometry.
Retaining masked locations therefore reveals the structure that the student should infer (see \cref{fig:augmented_views}).
Instead, as in object-level 3D methods~\citep{pang2022pointmae,leijenaar2025asymdsd}, we remove masked tokens entirely before the student encoder and use a lightweight cross-attention predictor to reconstruct their representations from visible context.
We supervise these predictions with the same targets introduced above, providing a unified supervision signal across both objectives.

Beyond self-distillation, knowledge distillation from pretrained 2D VFMs has become an important source of supervision for 3D representation learning~\citep{knaebel2025ditr,zhang2025concerto,zhang2026utonia}.
Current approaches train the 3D encoder to regress dense image features.
However, the same 3D region can yield different image embeddings across viewpoints, so regressing these features exposes the 3D encoder to inconsistent targets and encourages it to learn view-specific variation rather than a consistent semantic representation.
We instead construct multi-view-consistent targets in the frozen 2D feature space and train the 3D encoder to predict them.
This removes view-specific variation from the supervision signal and further improves representation quality (see \cref{fig:vfm_clustering}).

Together, these contributions establish \ours{} as a strong framework for 3D representation learning.
With 3D-only pretraining, \ours{} substantially improves linear probing over prior work by $+5.2$ mIoU on ScanNet200~\citep{rozenberszki2022scannet200} and $+3.5$ on ScanNet++~\citep{yeshwanthliu2023scannetpp}.
Incorporating 2D-to-3D distillation yields further improvements of $+4.5$ on ScanNet200, $+5.8$ on ScanNet++, and $+3.9$ on nuScenes~\citep{caesar2020nuscenes}, achieving the best linear-probing results across six benchmarks.
Beyond scene-level semantic segmentation, the frozen representations also transfer to zero-shot point-cloud registration and single-view RGB-D semantic segmentation under linear probing, demonstrating their versatility across downstream tasks.
\section{Related Work}

\PARbegin{Self-Distillation in 2D.}
BYOL~\citep{grill2020byol} demonstrated that strong visual representations can be learned by training a student network to predict the representations of an EMA teacher across differently augmented views of the same image.
DINO~\citep{caron2021dino} extends this paradigm by matching teacher and student probability distributions over learned prototypes using cross-entropy loss and adopting a multi-crop strategy~\citep{caron2020swav}, where the teacher processes global views while the student additionally processes local views.
iBOT~\citep{zhou2021ibot} augments self-distillation with masked image modeling, where the student additionally processes masked global views and predicts the teacher's prototype assignments at masked locations using the visible context.
DINOv2~\citep{oquab2023dinov2} and, more recently, DINOv3~\citep{simeoni2025dinov3} build on these objectives, substantially scaling training data and model capacity to learn general-purpose visual representations.
Alternatively, I-JEPA~\citep{assran2023ijepa} and CAPI~\citep{darcet2025capi} rely exclusively on masked image modeling without an additional global self-distillation objective.
While I-JEPA trains the student to regress teacher features directly, CAPI instead uses cross-entropy loss to predict balanced prototype assignments obtained by clustering teacher features, improving training stability without global supervision.

\PARbegin{SSL on 3D Scenes.}
Early 3D SSL approaches rely on contrastive objectives that enforce consistency across corresponding points, regions, or augmented scene views~\citep{xie2020pointcontrast,hou2021csc,zhang2021depthcontrast,chen20224dcontrast,wang2024groupcontrast}.
Subsequent methods complement contrastive learning with reconstruction objectives, such as recovering masked geometry, color, or surface normals~\citep{wu2023msc,xu2023mm3dscene}.
Inspired by 2D self-distillation~\citep{caron2021dino,zhou2021ibot,oquab2023dinov2}, Sonata~\citep{wu2025sonata} and MSM~\citep{hermosilla2025msm} concurrently extend teacher--student frameworks to 3D scenes.
Sonata trains a student on local and masked views to match teacher prototype distributions from global views, while MSM reconstructs teacher representations at masked locations through a hierarchical decoder.
Follow-up works extend the 3D self-distillation framework in complementary directions: DOS~\citep{abdelsamad2026dos} restricts supervision to observable regions and introduces long-tail-aware prototype assignments, PointINS~\citep{yang2026pointins} combines self-distillation with instance-aware geometric learning, and LAM3C~\citep{yamada2026lam3c} scales pretraining using video-generated point clouds.
In this work, we identify limitations of these current approaches and revisit their core design choices, resulting in substantially stronger representations.

\PARbegin{2D--3D Distillation.}
A parallel line of work distills pretrained VFMs into 3D encoders for representation learning~\citep{puy2024three,kong2025largead}. DITR~\citep{knaebel2025ditr} distills dense 2D foundation-model features into a 3D backbone, while Concerto~\citep{zhang2025concerto} combines 3D self-distillation with cross-modal 2D--3D alignment. Utonia~\citep{zhang2026utonia} further extends this paradigm toward joint pretraining across heterogeneous point-cloud domains.
These approaches demonstrate that pretrained image representations provide powerful semantic supervision for 3D representation learning, but overlook the view-dependent information encoded in VFM features.
Instead, we construct a view-consistent target space before transferring this supervision to 3D.
\section{Method}
\begin{figure}[t]
    \centering
    {
    \let\fsize\relax
\includegraphics[trim=0bp 15bp 0bp 0bp, clip,
                   width=\linewidth]{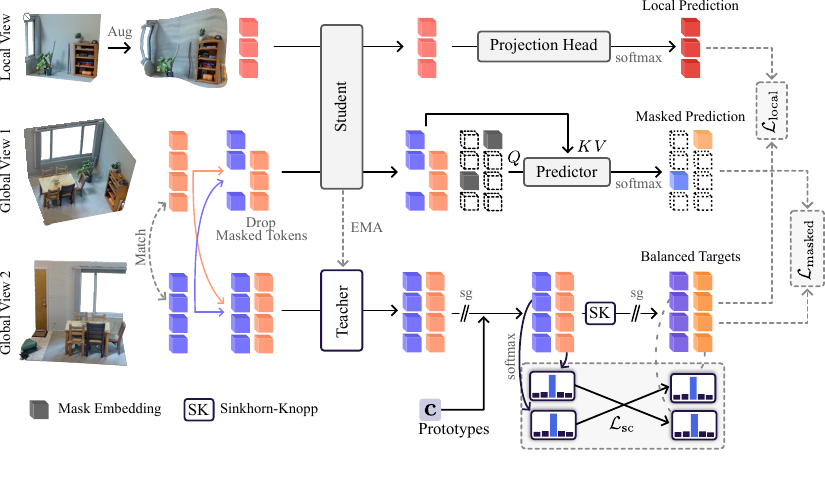}
    }
    \vspace{-10pt}
\caption{\textbf{\ours{} Overview.} Overlapping global and local views are sampled from a 3D scene. The EMA teacher encodes clean global views and softly assigns their tokens to learned prototypes, producing balanced targets. The prototypes are optimized independently by encouraging balanced and consistent assignments across global views. The student encodes strongly augmented local views and masked global views. Local-view embeddings, through a projection head, predict the targets of corresponding global-view tokens. Masked view embeddings provide context to a cross-attention-only predictor that predicts the targets at masked locations.
    }
    \label{fig:main_method}
\end{figure}

We revisit four key design choices of 3D self-distillation frameworks: the encoder, local prediction, masked prediction, and 2D--3D distillation.
\cref{fig:main_method} provides an overview.
We first establish a stronger baseline by replacing the Point Transformer V3 (PTv3)~\citep{wu2024ptv3} encoder of Sonata~\citep{wu2025sonata} with the Volume Transformer (Volt)~\citep{yilmaz2026volume} (\cref{sec:baseline}).
For local prediction, we feed clean global views to the teacher and strongly augmented local views to the student, and decouple the teacher's target head from the student's prediction head (\cref{sec:local-prediction}).
For masked prediction, we remove masked tokens from the student encoder to avoid any geometric shortcut, and predict their targets solely from the visible context with a cross-attentive predictor (\cref{sec:masked-prediction}).
Finally, in a second training stage, we distill a frozen 2D vision foundation model (VFM) into the 3D encoder using view-consistent targets derived from its features (\cref{sec:2d3d-distillation}).

\subsection{Baseline}
\label{sec:baseline}
We start from Sonata~\citep{wu2025sonata}, an established 3D self-distillation framework in which a student is trained to match the soft assignments of an EMA teacher over a set of learned prototypes~\citep{caron2021dino}.
From each input scene, Sonata samples several overlapping global and local views.
The teacher processes the global views to produce targets, and the student is trained on two objectives.
In \emph{local prediction}, the student encodes the local views and predicts the teacher's assignments at the corresponding locations in the global views.
In \emph{masked prediction}, the student encodes masked copies of the global views and predicts the teacher's assignments for the masked voxels.
Unlike 2D methods, which typically apply local prediction only to a global class token, Sonata applies both objectives densely.
Its backbone is the hierarchical PTv3 encoder~\citep{wu2024ptv3}, and the objectives are applied to features obtained by upsampling and concatenating the outputs of several encoder stages.

We first replace PTv3 with Volt~\citep{yilmaz2026volume}, currently the best-performing 3D backbone.
Since Volt operates on a single-resolution sequence of volumetric tokens, dense supervision can be applied directly to its output tokens, without a need for multi-stage upsampling and concatenation.
{\parfillskip=0pt\par}
\begin{wraptable}[9]{r}{0.32\textwidth}
\small
\vspace{-1pt}
\centering
\begin{tabular*}{\linewidth}{@{\extracolsep{\fill}}lc@{}}
    \toprule
& \textbf{ScanNet200} \\
    \midrule
    Sonata\textsuperscript{\(\dagger\)} (PTv3)  & $28.3$ \\
    Sonata (Volt)  & $\mathbf{29.6}$ \\
    Simpler scheduling & $29.5$ \\
\bottomrule
\end{tabular*}
\vspace{-0pt}
\caption{\small\textbf{Baseline.}
Semantic segmentation linear-probing performance (mIoU).
\textsuperscript{\(\dagger\)} is reproduced.}
\label{tab:baseline}
\end{wraptable}
\noindent This change alone improves ScanNet200 linear probing by 1.3 mIoU (\cref{tab:baseline}).
Volt also trains more stably, which lets us simplify the hyperparameter scheduling.
Following DINOv3~\citep{simeoni2025dinov3}, we keep the learning rate constant after warmup, fix the weight decay, and drop all other schedules.
This matches the linear probing performance of the original setup while eliminating its dependence on a predefined training length, so training can continue for as long as performance improves.

\subsection{Local Prediction}
\label{sec:local-prediction}
Removing the masked prediction branch from our Volt baseline lowers ScanNet200 linear probing by only 0.6 mIoU (\cref{tab:local}), indicating that training is driven primarily by local prediction.
We therefore first study local prediction in isolation and revisit masked prediction later in \cref{sec:masked-prediction}.

\begin{wraptable}[9]{r}{0.32\textwidth}
\small
\vspace{-10pt}
\centering
\begin{tabular*}{\linewidth}{@{\extracolsep{\fill}}lc@{}}
    \toprule
    & \textbf{ScanNet200} \\
    \midrule
    No masked loss           & $28.9$ \\
    Asymmetric augm.         & $31.3$ \\
    Decoupled head           & $\mathbf{32.8}$ \\
    \bottomrule
\end{tabular*}
\vspace{0pt}
\caption{\small\textbf{Local Prediction.}
The masked branch is removed. Same evaluation setup as in \cref{tab:baseline}.}
\label{tab:local}
\end{wraptable}

\PARbegin{Asymmetric Views.}
Sonata applies the same set of augmentations to the global teacher views and the local student views (\cref{fig:augmented_views}).
This exposes the teacher to the same strong geometric perturbations as the student, which can degrade the targets it produces.
We instead keep the teacher's global views clean and apply strong corruptions, such as elastic distortion and point dropout, only to the student's local views.
As a result, the student must recover the teacher's assignments from substantially corrupted inputs.
This change alone improves linear probing by 2.4 mIoU.

\PARbegin{Decoupled Target Head.}
The asymmetry also changes how the teacher should construct its targets.
In Sonata, the teacher's target head is an EMA copy of the student's prediction head.
With asymmetric views, however, the two heads receive features from different input distributions: the student head from corrupted local views, the teacher head from clean global views.
Through the EMA, the teacher head would track a head adapted to corrupted inputs and could not specialize to clean ones.
We therefore decouple them and equip the teacher with an independent target head, described next.

\PARbegin{Teacher Target Construction.}
The teacher head maps teacher encoder outputs to soft assignments over $K$ prototypes, which serve as targets for the student.
Inspired by CAPI~\citep{darcet2025capi}, we parameterize it as a set of learned prototypes $\{c_k\}_{k=1}^{K}$ with $c_k \in \mathbb{R}^{D}$, independent of the student head.
Given the teacher's encoder outputs $\{x_i\}_{i=1}^{N}$ with $x_i \in \mathbb{R}^{D}$, we compute their cosine similarities $\ell \in \mathbb{R}^{N \times K}$ to the prototypes, sharpen them with a temperature $\tau_t$, and apply the Sinkhorn--Knopp (SK) algorithm~\citep{sinkhorn1967sinkhornknopp} to obtain soft assignments $P_t \in \mathbb{R}^{N \times K}$ that are balanced across prototypes:
\begin{equation}
\ell_{i,k}
=
\frac{c_k^\top x_i}
{\lVert c_k\rVert_2 \lVert x_i\rVert_2},
\qquad
P_t
=
\operatorname{SK}\!\left(\frac{\ell}{\tau_t}\right).
\label{eq:teacher-assignments}
\end{equation}
The student's prediction head outputs a $K$-way distribution for each student token and is trained with a cross-entropy loss $\mathcal{L}_{\mathrm{local}}$ against the teacher assignment $P_t$ at the corresponding location in a global view.
Decoupling target construction from student prediction lets the teacher fit its target space to clean encoder representations, while the student's prediction head is free to adapt to the corrupted inputs from which these targets must be recovered.

\PARbegin{Learning the Prototypes.}
The self-distillation objective trains the student, but it cannot also update the teacher prototypes, as this would couple the targets to the predictions it supervises and lead to collapse.
We therefore learn the prototypes independently using correspondences between the two global teacher views.
Let $\mathcal{M}$ denote the set of pairs $(i,j)$ of corresponding teacher tokens from the two views.
For each pair, the balanced assignment of one token, which also supervises the student, serves as a fixed target for the other token's distribution over prototypes, and vice versa:
\begin{equation}
\mathcal{L}_{\mathrm{sc}}
=
\frac{1}{2\,|\mathcal{M}|}
\sum_{(i,j)\in\mathcal{M}}
\left[
\operatorname{CE}\!\left(P_{t,i}, q_j\right)
+
\operatorname{CE}\!\left(P_{t,j}, q_i\right)
\right],
\qquad
q_i
=
\operatorname{softmax}\!\left(\frac{\ell_i}{\tau_C}\right),
\label{eq:swapped-clustering}
\end{equation}
where $\tau_C$ is a temperature.
We stop gradients through the teacher encoder and the Sinkhorn-Knopp assignments~\citep{asano2019sela}, so this swapped clustering loss~\citep{caron2020swav} updates only the prototypes.
Together, the decoupled target head and its independently learned prototypes further improve linear probing by 1.5 mIoU (\cref{tab:local}).

\subsection{Masked Prediction}
\label{sec:masked-prediction}
In masked prediction, the student must infer the teacher's targets for hidden regions from the visible context.
In 2D vision, masking is typically implemented by replacing the selected tokens with a learned mask embedding while keeping them in the sequence, and existing 3D approaches carry this strategy over to point clouds~\citep{wu2023msc,wu2025sonata,zhang2025concerto,zhang2026utonia}.
In sparse 3D data, however, occupancy itself carries geometric information.
Mask tokens kept at their original 3D positions therefore reveal the shape of the masked region.
This creates a \emph{geometric shortcut}~\citep{pang2022pointmae,knaebel2023point2vec,hermosilla2025msm}: the student can predict the targets from the geometry of the masked region itself rather than from the surrounding visible context.

\begin{wraptable}[8]{r}{0.31\textwidth}
\small
\centering
\vspace{-10px}
\begin{tabular*}{\linewidth}{@{\extracolsep{\fill}}lc@{}}
    \toprule
    & \textbf{ScanNet200} \\
    \midrule
    Local pred.            & $32.8$ \\
    + Masked pred.         & $35.6$ \\
    + Fix artifacts        & $\mathbf{35.9}$ \\
    \bottomrule
\end{tabular*}
\vspace{-5px}
\caption{\textbf{Masked Prediction.} Same evaluation setup as in \cref{tab:baseline}.
}
\label{tab:masked}
\end{wraptable}
\PARbegin{Cross-Attentive Predictor.}
We remove this shortcut by dropping the masked tokens before the student encoder, which then processes only the visible tokens.
To obtain predictions at the masked locations, we attach a lightweight Transformer decoder consisting solely of cross-attention layers~\citep{fu2024crossmae,leijenaar2025asymdsd}.
For each masked token, the decoder initializes a query with a learned mask embedding and places it at the 3D position of that token.
In every decoder layer, the queries cross-attend to the encoded visible tokens, using the same 3D positional encoding as the encoder.
We deliberately omit self-attention among the queries.
Each query therefore sees only its own position and the visible context, and cannot infer the geometry of the masked region from the positions of other masked queries.
The decoder outputs are supervised with targets that the teacher computes at the same locations from the complete, unmasked global view.

\PARbegin{Shared Target Space.}
As in local prediction, the student and teacher operate on different features: the student head receives decoder outputs from a masked view, while the teacher constructs targets from encoder outputs on the complete view.
The decoupled target head is therefore equally suited to this setting, and we reuse the teacher assignments $P_t$ from \cref{eq:teacher-assignments} as targets.
A dedicated student prediction head, separate from the one used for local prediction, maps each decoder output to a $K$-way distribution, which is trained with a cross-entropy loss against the teacher assignment at the same location.
Adding this masked objective back improves ScanNet200 linear probing by 2.8 mIoU (\cref{tab:masked}).
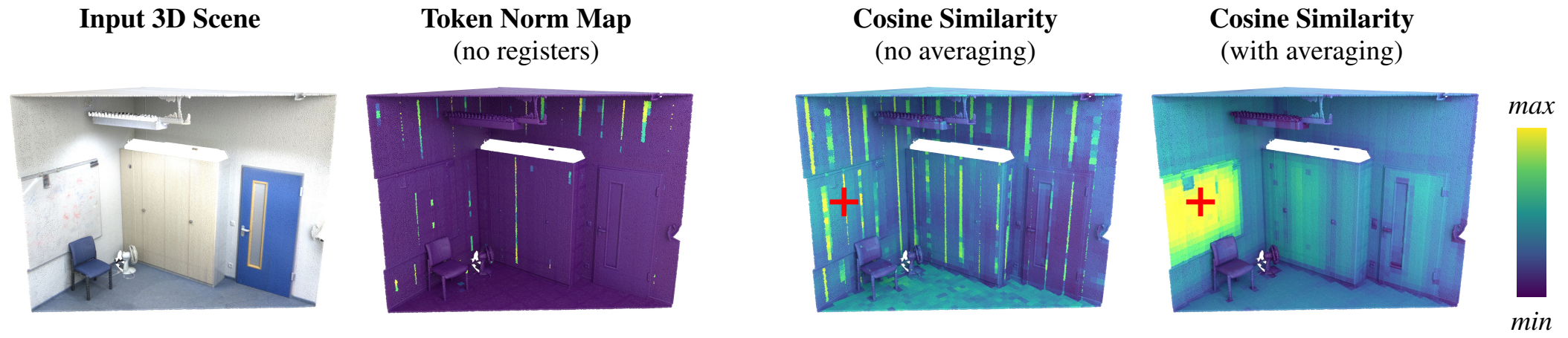
\begin{figure}[t]
\centering
    \newcommand{\artifactpanel}[4]{\begin{minipage}[t]{0.22\linewidth}
            \centering
            \parbox[t][1.6\baselineskip][t]{\linewidth}{\centering\scriptsize
                \textbf{#1}#2}\par\vspace{-1pt}
            \begin{tikzpicture}[baseline=(panel.south)]
                \node[inner sep=0] (panel) {
                    \includegraphics[
                        trim={0 0 0 65px},
                        clip,
                        width=\linewidth
                    ]{#3}
                };

\fill[white,overlay]
                    ([xshift=0.13\linewidth,yshift=0\linewidth]panel.north west) --
                    ([xshift=0.42\linewidth,yshift=0\linewidth]panel.north west) --
                    ([xshift=0.42\linewidth,yshift=-0.02863\linewidth]panel.north west) --
                    ([xshift=0.13\linewidth,yshift=-0.04637\linewidth]panel.north west) -- cycle;

                #4
            \end{tikzpicture}
        \end{minipage}}

    \artifactpanel
        {Input 3D Scene}
        {}
        {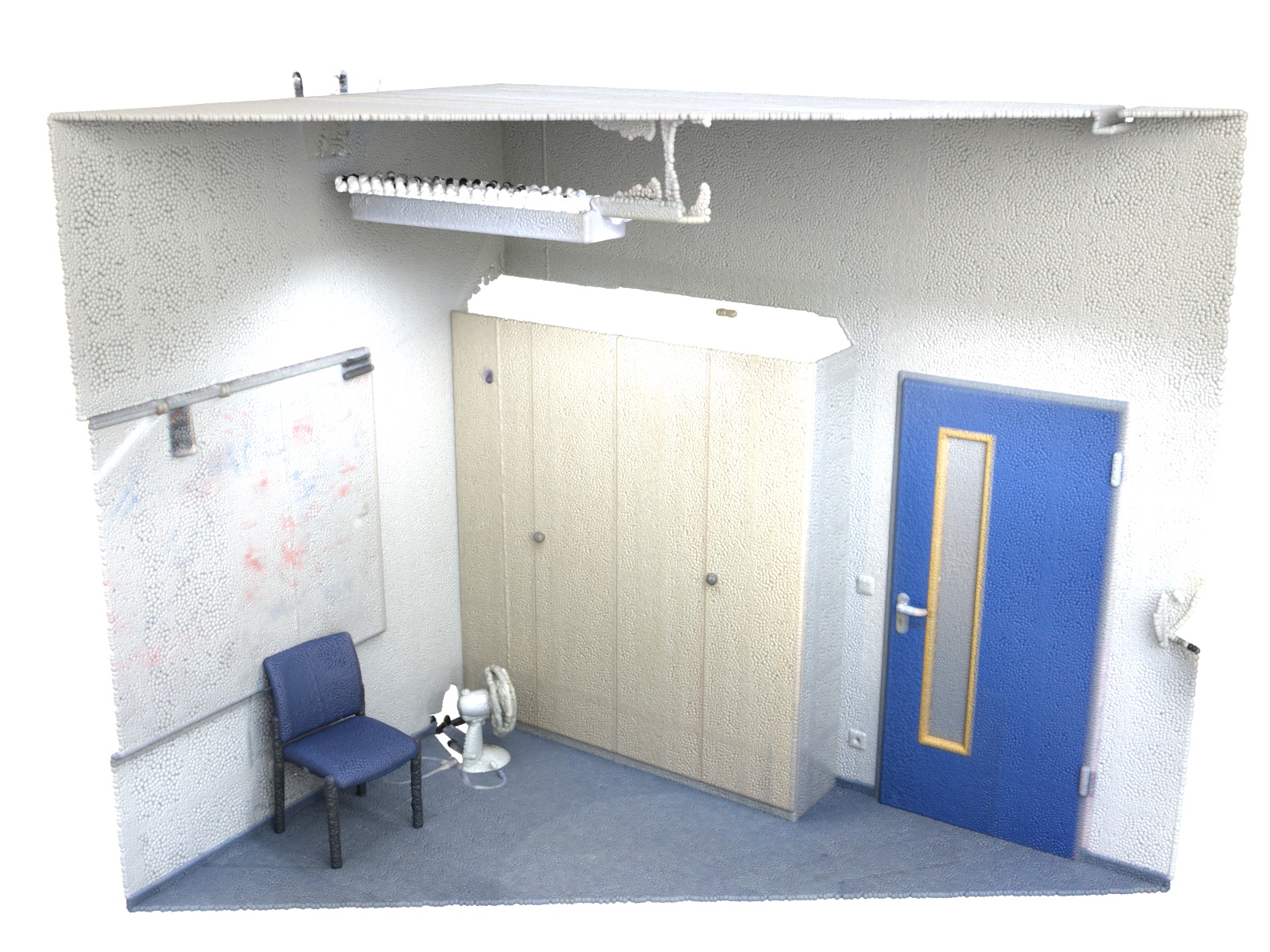}
        {}
    \hfill
    \artifactpanel
        {Token Norm Map}
        {\\\normalfont (no registers)}
        {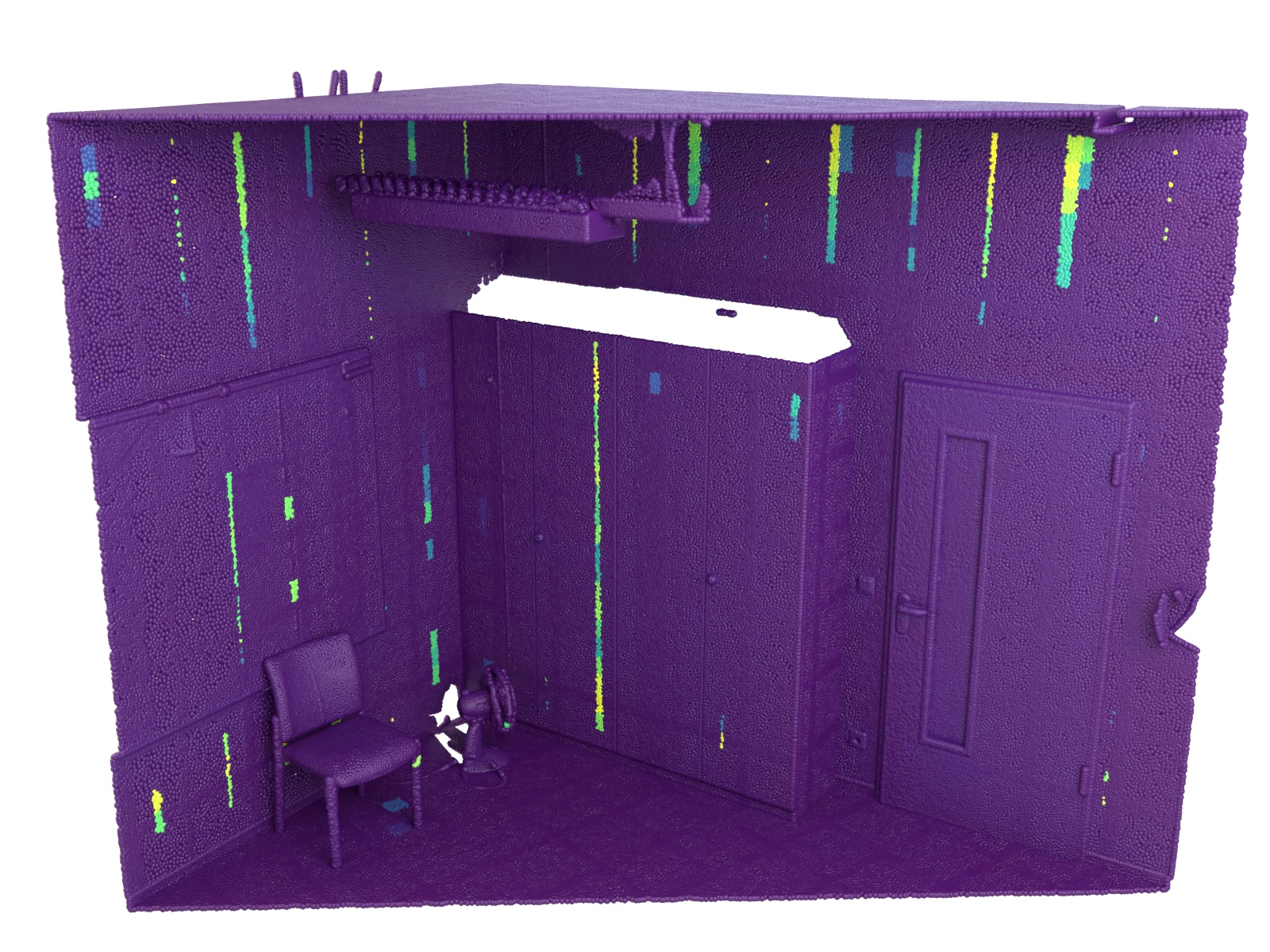}
        {}
    \hfill
    \hspace{0.04\linewidth} \artifactpanel
        {Cosine Similarity}
        {\\\normalfont (no averaging)}
        {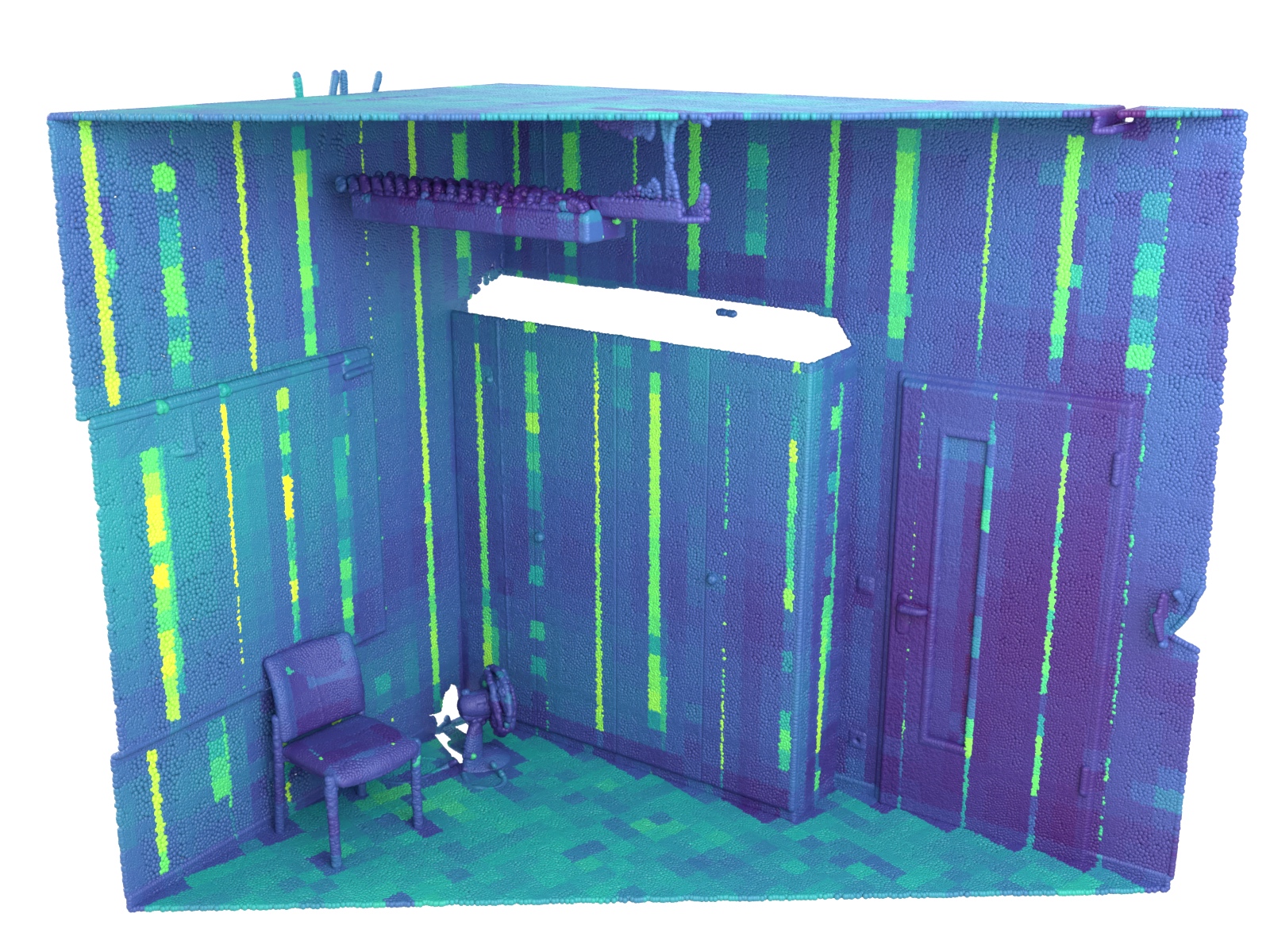}
        {
            \node[overlay,red,font=\Large\bfseries] at
                ([xshift=0.18\linewidth,yshift=0.3525\linewidth]panel.south west)
                {\textbf{+}};
        }
    \hfill
    \artifactpanel
        {Cosine Similarity}
        {\\\normalfont (with averaging)}
        {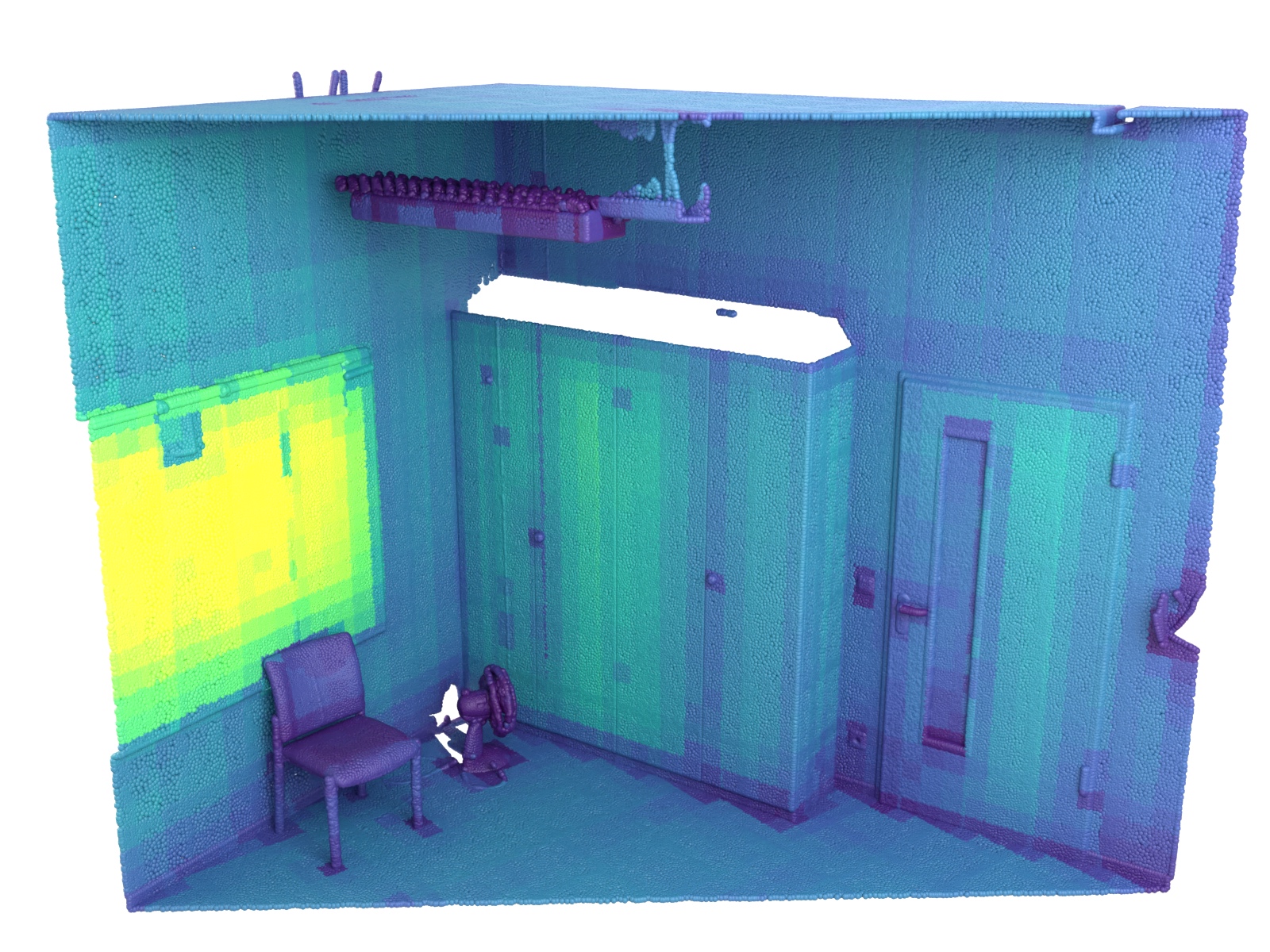}
        {
            \node[overlay,red,font=\Large\bfseries] at
                ([xshift=0.18\linewidth,yshift=0.3525\linewidth]panel.south west)
                {\textbf{+}};
        }
    \hfill
    \raisebox{-0.172\linewidth}{\begin{tikzpicture}[baseline=(legend.south)]
            \node[inner sep=0] (legend) {
                \includegraphics[
                    width=0.01864\linewidth,
                    height=0.10725\linewidth
                ]{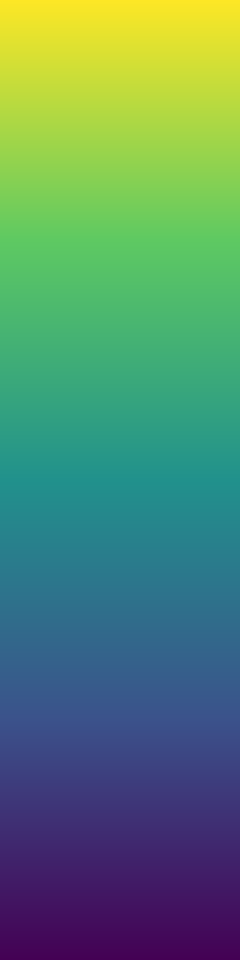}
            };
            \node[anchor=south,font=\scriptsize\itshape] at (legend.north) {max};
            \node[anchor=north,font=\scriptsize\itshape] at (legend.south) {min};
        \end{tikzpicture}}
    \vspace{-5pt}
    \caption{
        \textbf{Representation Artifacts.}
        We observe high-norm token artifacts in the representations (left). Register tokens largely eliminate these artifacts, but early in training, the learned representations still strongly reflect local voxel occupancy. Averaging voxel features within each patch removes this residual artifact (right).
}
    \label{fig:artifacts}
\end{figure}

\PARbegin{Preventing Representation Artifacts.}
During training, we observe high-norm token artifacts similar to those reported for ViTs~\citep{darcet2024registers}.
Adding learned register tokens~\citep{darcet2024registers} removes most of these outliers, but some persist, particularly early in training.
We trace these remaining artifacts to the tokenizer.
Volt tokenizer aggregates the voxel features within each patch with a weighted sum, so the magnitude of the resulting token depends on how many voxels in the patch are occupied.
Early in training, before semantic features emerge, the model can exploit these occupancy-dependent magnitudes as a shortcut for matching the teacher.
We remove this shortcut by normalizing the weighted sum by the number of occupied voxels, i.e., averaging rather than summing.
This decouples token magnitude from occupancy and substantially reduces the remaining artifacts (\cref{fig:artifacts}), while also improving ScanNet200 linear probing by 0.3 mIoU (\cref{tab:masked}).

\subsection{2D--3D Distillation}
\label{sec:2d3d-distillation}
Vision foundation models (VFMs) are trained on large and diverse image collections and provide semantic features that can complement those learned from 3D point clouds.
Recent 3D representation learning methods therefore supervise 3D encoders with VFM features extracted from aligned posed images~\citep{knaebel2025ditr,zhang2025concerto,zhang2026utonia}.
We inspect two aspects of this distillation: when VFM supervision should be introduced, and how VFM features should be converted into targets.

\PARbegin{Two-Stage Training.}
Experimentally, we find that applying 2D--3D distillation from the start of training underperforms first learning the 3D representation fully self-supervised and aligning it with the VFM afterwards, by 1.4 mIoU on ScanNet200 linear probing (\cref{tab:2d3ddistill}).
We therefore divide pretraining into two stages.
The first stage trains the encoder only with the self-supervised objectives of \cref{sec:local-prediction,sec:masked-prediction}.
The second stage continues from this checkpoint, keeps all self-supervised objectives active, and adds a distillation objective from a frozen VFM~\citep{oquab2023dinov2}.

\begin{wraptable}[9]{r}{0.31\textwidth}
\small
\centering
\vspace{-12px}
\begin{tabular*}{\linewidth}{@{\extracolsep{\fill}}lc@{}}
\toprule
    & \textbf{ScanNet200} \\
    \midrule
Align from start        & $37.2$ \\
    \midrule
    Raw embeddings          & $38.6$ \\
    Prototypes              & $\mathbf{40.4}$ \\
    \bottomrule
\end{tabular*}
\vspace{-5px}
\caption{\textbf{2D--3D Distillation.}
Linear probing (mIoU) for different 2D--3D distillation objectives.}
\label{tab:2d3ddistill}
\end{wraptable}

\PARbegin{View-Consistent Targets.}
Previous approaches typically project 3D features linearly into the VFM embedding space and align them with the corresponding 2D features using a cosine-similarity objective~\citep{knaebel2025ditr,zhang2025concerto}.
However, VFM features encode not only semantic content but also view-dependent information, such as image-grid position~\citep{cuttano2026insid3}.
Directly regressing these features can therefore assign different targets to the same 3D location when it is observed from different viewpoints, which yields inconsistent supervision for a view-independent 3D encoder.

To this end, we introduce a separate set of learned prototypes for the VFM features and softly assign them to the 2D tokens via cosine similarities and Sinkhorn--Knopp, as done for the 3D teacher (\cref{eq:teacher-assignments}).
Using camera poses and depth, we identify pairs of 2D tokens in overlapping images that observe the same surface location.
We train the prototypes on these pairs with the same swapped clustering loss $\mathcal{L}_{\mathrm{sc}}$ (\cref{eq:swapped-clustering}), keeping the VFM frozen.
Because each token's assignment must predict that of its counterpart in the other view, view-specific components of the VFM features do not help reduce this loss.
The learned prototypes therefore capture the view-consistent part of the VFM representation and suppress view-dependent variation (\cref{fig:vfm_clustering}).

To supervise the 3D encoder, we project the point cloud into the images and average the encoded 3D features of all points that fall into the same 2D token.
A dedicated prediction head on the student then predicts that 2D token's soft prototype assignment and is trained with a cross-entropy loss.
Compared with directly regressing VFM features, this improves ScanNet200 linear probing by 1.8 mIoU (\cref{tab:2d3ddistill}).

\newcommand{\boxedimage}[5]{\begin{tikzpicture}
    \node[anchor=south west, inner sep=0] (img) at (0,0)
        {\includegraphics[width=#2]{#1}};
    \begin{scope}[x={(img.south east)}, y={(img.north west)}]
        \draw[#5, very thick] (#3) rectangle (#4);
    \end{scope}
\end{tikzpicture}}

\begin{figure}[t]
    \centering

    \begin{minipage}[t]{0.31\textwidth}
        \centering
        \small\textbf{Multi-view Images}\strut\\[3pt]
            \boxedimage{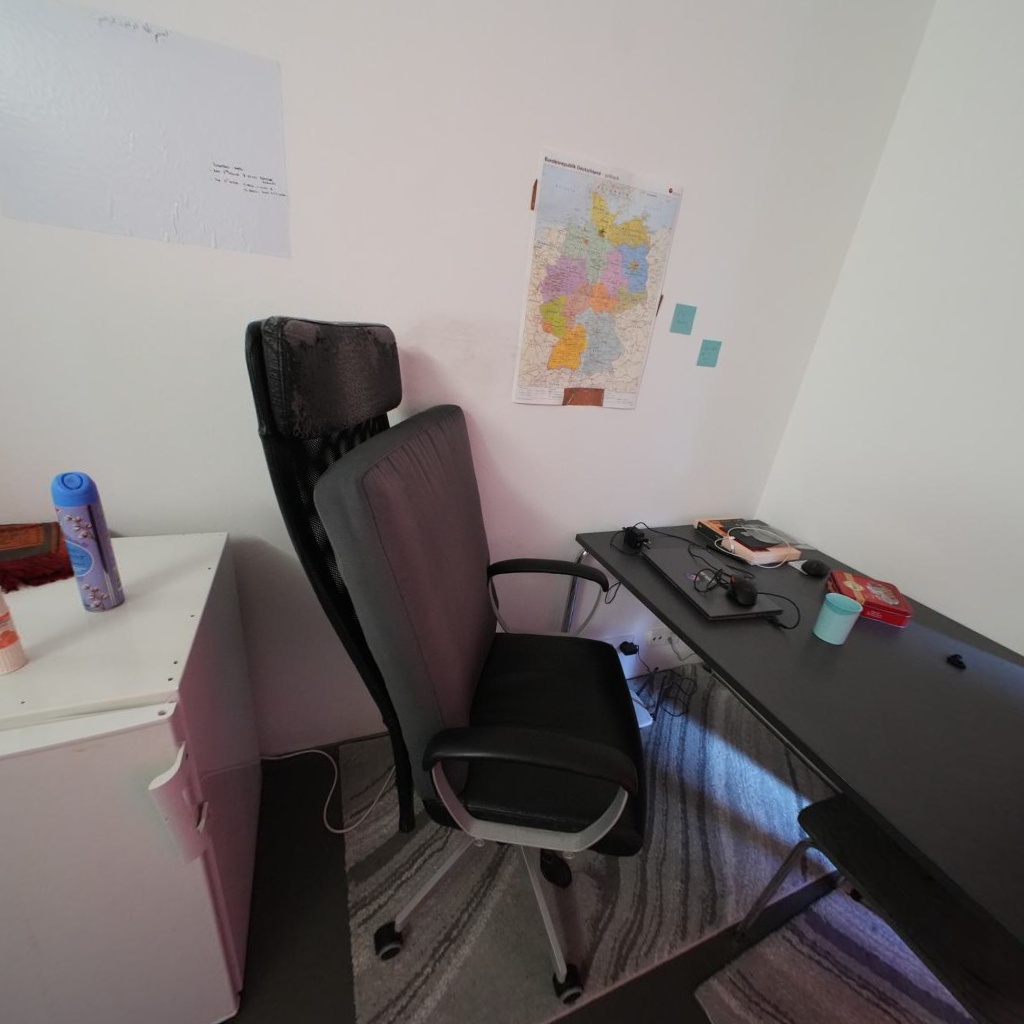}{0.47\linewidth}{0.01,0.37}{0.21,0.57}{red}
\hspace{0.01\linewidth}
        \boxedimage{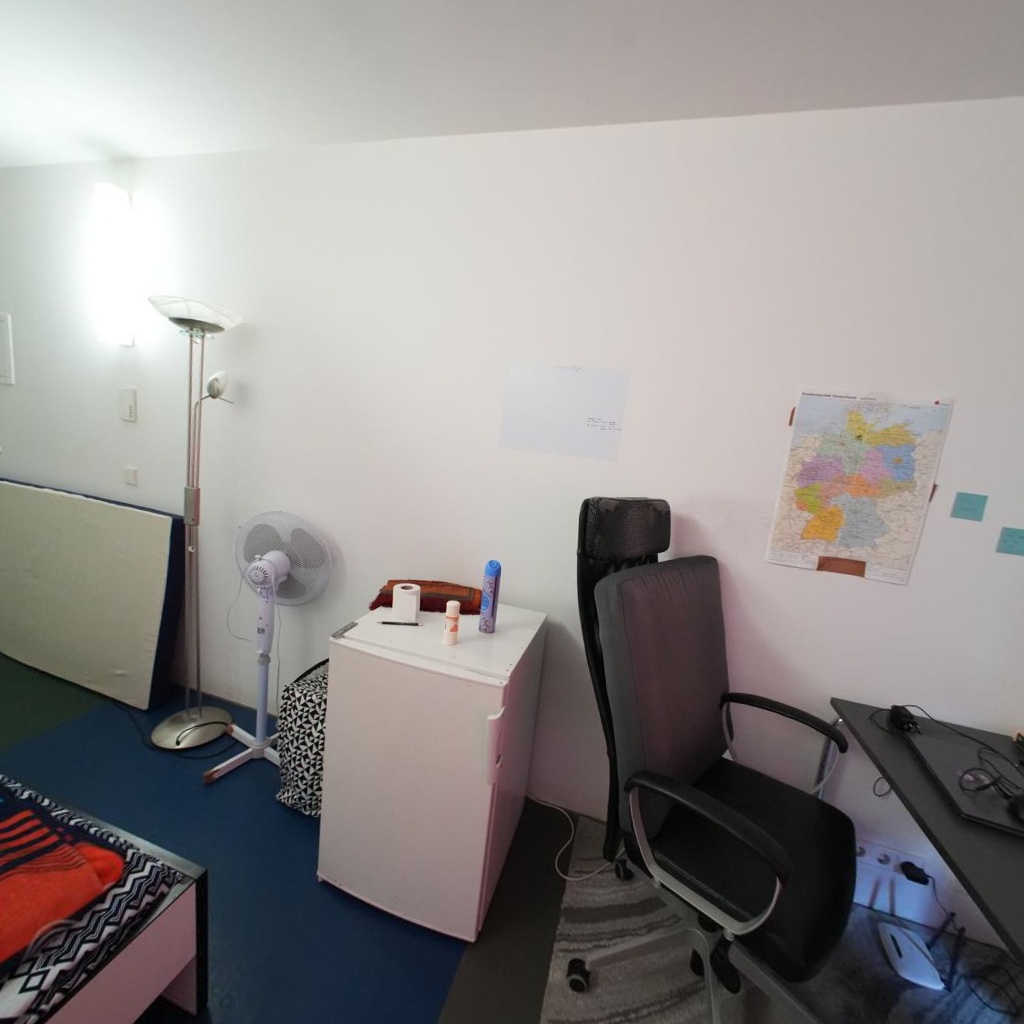}{0.47\linewidth}{0.35,0.3}{0.55,0.5}{red}
\end{minipage}
    \hfill
    \begin{minipage}[t]{0.31\textwidth}
        \centering
        \small\textbf{VFM Features (DINOv2)}\strut\\[3pt]
        \boxedimage{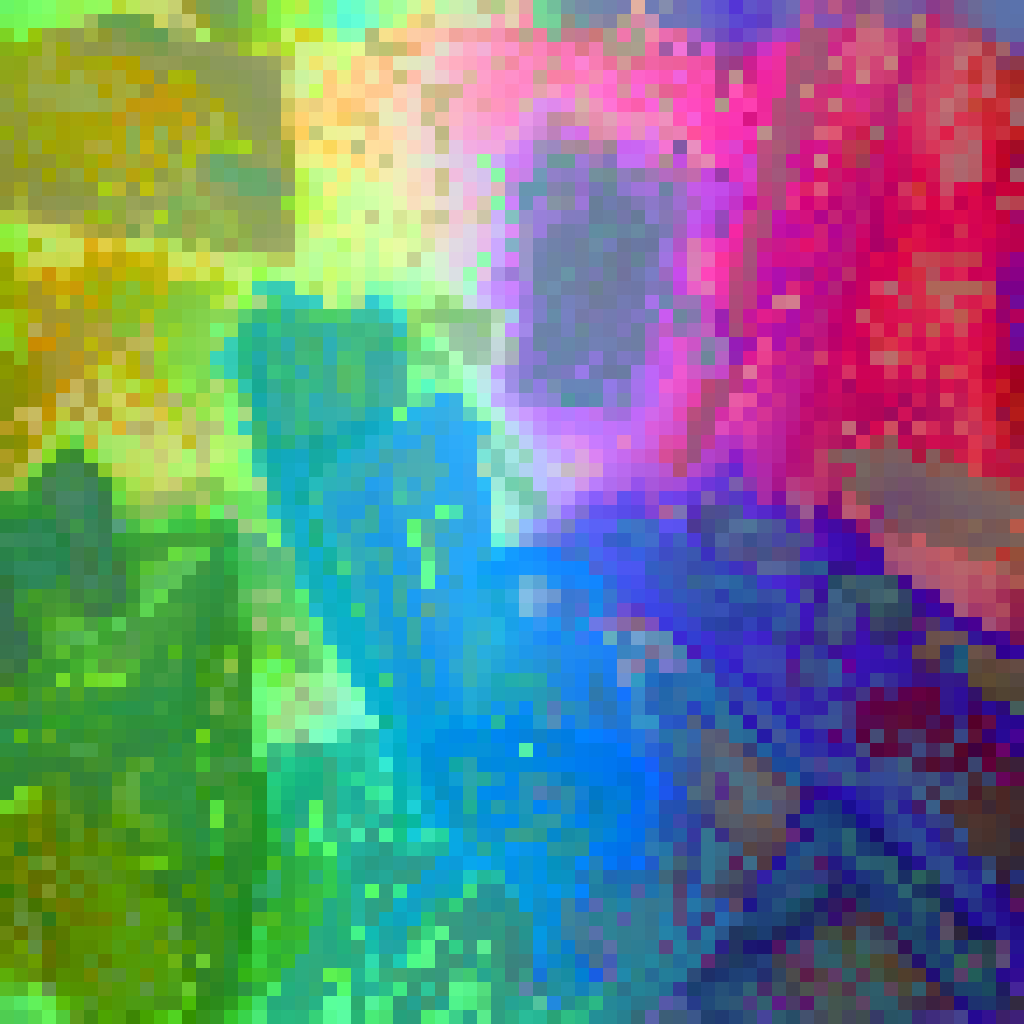}{0.47\linewidth}{0.01,0.37}{0.21,0.57}{red}
\hspace{0.01\linewidth}
        \boxedimage{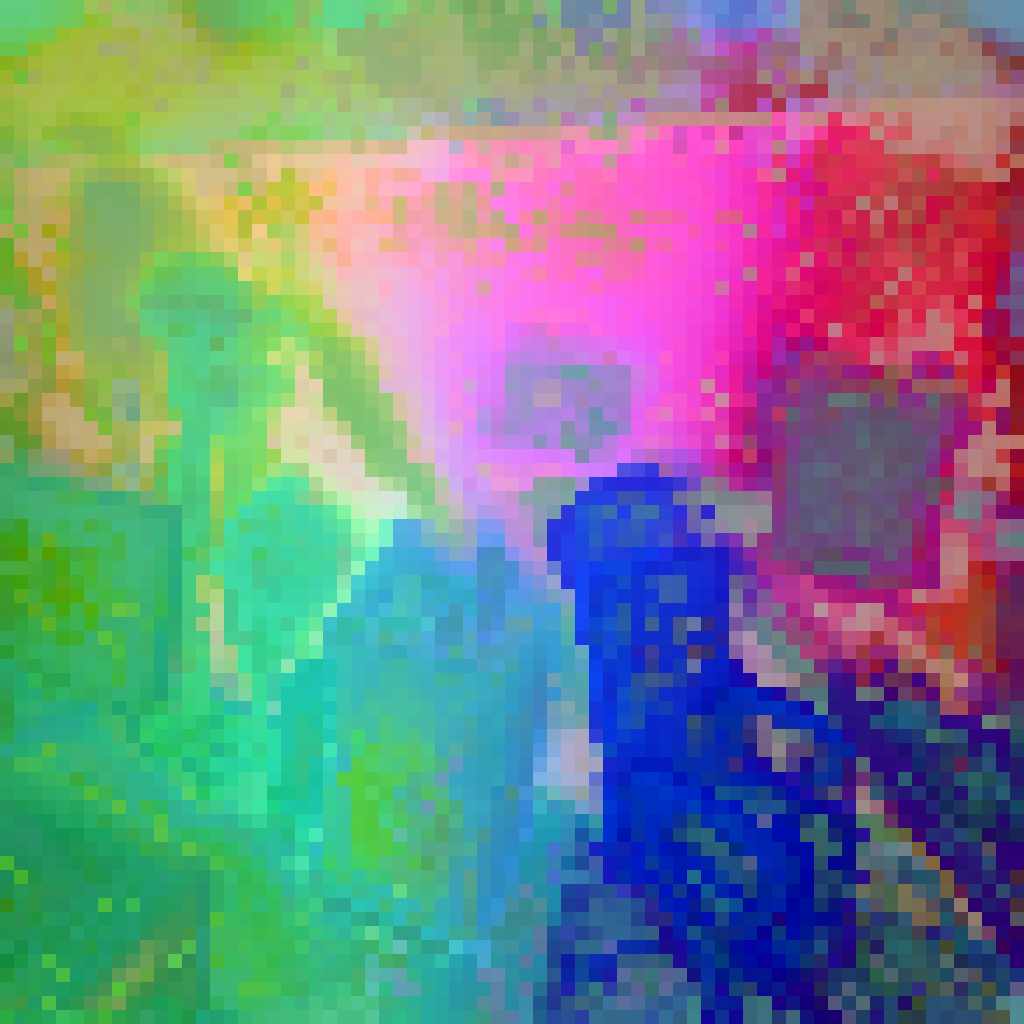}{0.47\linewidth}{0.35,0.3}{0.55,0.5}{red}
\end{minipage}
    \hfill
    \begin{minipage}[t]{0.31\textwidth}
        \centering
        \small\textbf{Consistent Targets (Ours)}\strut\\[3pt]
        \boxedimage{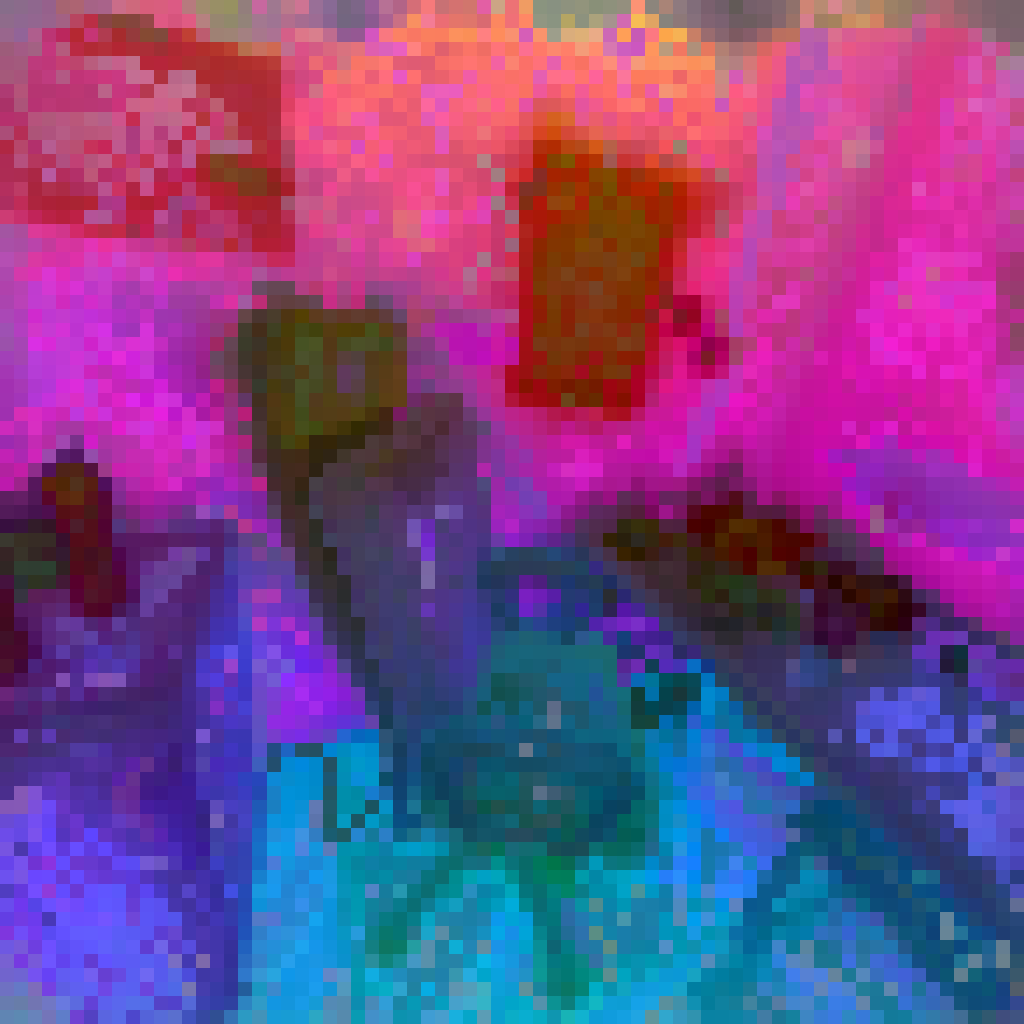}{0.47\linewidth}{0.01,0.37}{0.21,0.57}{red}
\hspace{0.01\linewidth}
        \boxedimage{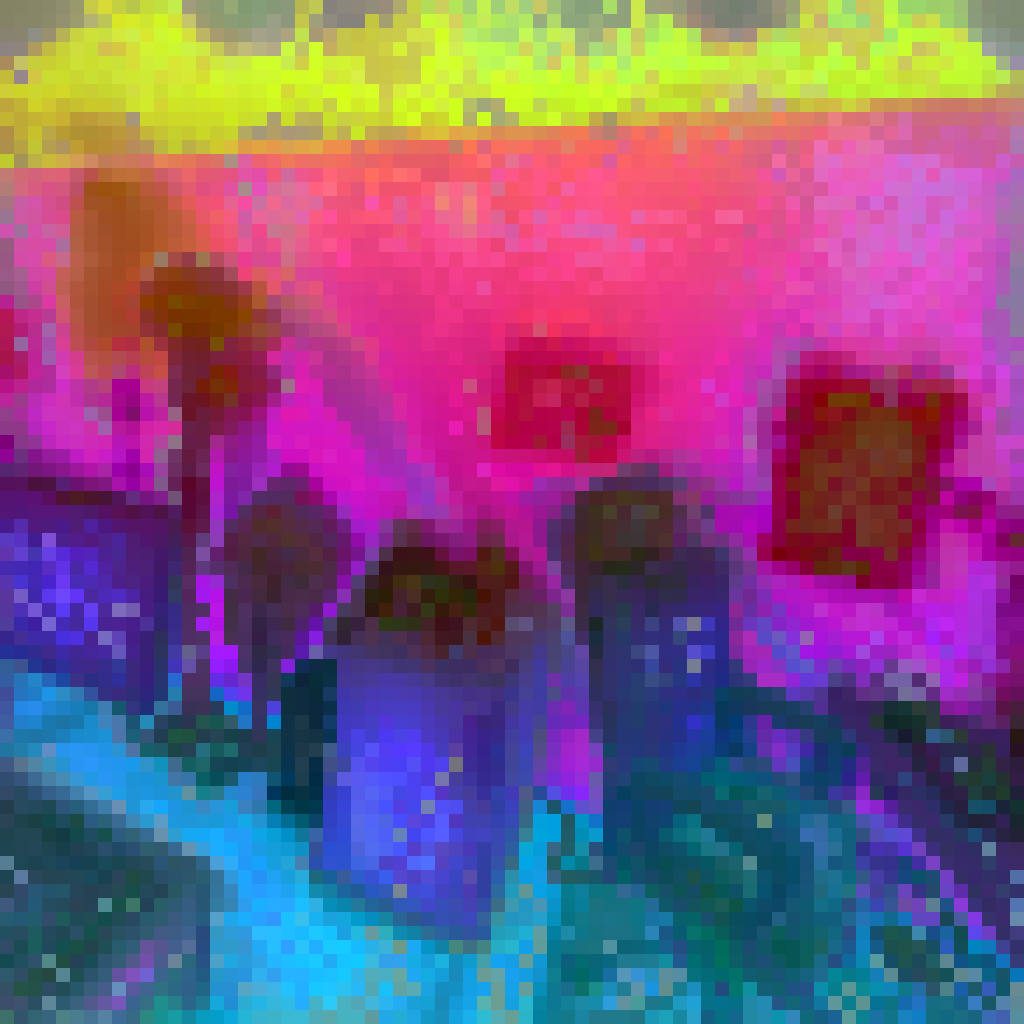}{0.47\linewidth}{0.35,0.3}{0.55,0.5}{red}
\end{minipage}
    \vspace{-5pt}
    \caption{
        \textbf{Consistent 2D Targets.}
VFM~\citep{oquab2023dinov2} features of the same 3D region vary across views. We learn prototypes in the frozen VFM feature space, optimized such that corresponding tokens across views produce matching soft assignments. These serve as our targets for 2D--3D distillation.
    }
    \label{fig:vfm_clustering}
    \vspace{-10pt}
\end{figure}

\PARbegin{Robustness.}
To improve robustness across diverse point clouds, where color may be unavailable and scenes may not be gravity-aligned, we extend the local-view augmentations in the second training stage with random color dropout and unrestricted $\mathrm{SO}(3)$ rotations.
Because these augmentations are applied only to the student's local views, the student must recover targets computed from clean and colored global views, which encourages invariance to both missing color and arbitrary orientation.
We introduce them only in the second stage, as these invariances matter for downstream use but are not needed for learning representations from 3D geometry in the first stage.
\section{Experiments}

\PARbegin{Pretraining Data.}
For indoor scenes, the 3D-only first stage uses the real-scene subset of Sonata's pretraining data~\citep{wu2025sonata}, comprising ScanNet~\citep{dai2017scannet}, ScanNet++~\citep{yeshwanthliu2023scannetpp}, S3DIS~\citep{armeni2016s3dis}, ARKitScenes~\citep{baruch2021arkitscenes}, and HM3D~\citep{ramakrishnan2021hm3d}, totaling approximately 18,000 scenes.
The second stage additionally requires posed RGB images for 2D--3D distillation.
As in prior work~\citep{zhang2025concerto,zhang2026utonia}, we further include RealEstate10K~\citep{zhou2018re10k}, increasing the indoor pretraining set to approximately 90,000 scenes.
We reconstruct its video sequences into point clouds using Depth Anything~3~\citep{lin2025depthanything3}.
For outdoor scenes, both stages use approximately 110,000 LiDAR scans from nuScenes~\citep{caesar2020nuscenes}, SemanticKITTI~\citep{behley2019semantickitti}, and Waymo~\citep{sun2020waymo}, with the second stage pairing them with synchronized camera images.

\PARbegin{Implementation Details.}
We train separate models for indoor and outdoor scenes, each using a Volt-B encoder with 16 register tokens.
In the first stage, following Sonata~\citep{wu2025sonata}, we sample two global crops with ratios of 0.4--1.0 and four local crops with ratios of 0.1--0.4 from each scene.
We block-mask the student's global views with a masking ratio of 55\% and a block size of 8~tokens (80\,cm indoors and 2\,m outdoors).
We train with AdamW for 200k steps at a batch size of 32.
In the second stage, we initialize from the first-stage checkpoint and train for a further 30k steps, using a frozen DINOv2-g/14~\citep{oquab2023dinov2} image encoder as the 2D teacher, following prior work~\citep{zhang2025concerto,zhang2026utonia}.
Additional details are provided in \cref{app:implementation}.

\begin{table*}[t]
\centering
\notfootnotesize
\setlength{\tabcolsep}{1pt}
\renewcommand{\probingbest}[1]{\probingscore{probingbestbg}{\boldmath\textbf{#1}}}
\renewcommand{\arraystretch}{1.08}

\begin{tabularx}{\textwidth}{c>{\hspace{3pt}}l>{\hspace{1pt}}lYYYYYY}
\toprule
 & \multirow{2}[2]{*}{\textbf{Method}} & \multirow{2}[2]{*}{\textbf{Backbone}} & \multicolumn{3}{c}{\textbf{{Indoor Datasets}}}
        & \multicolumn{3}{c}{\textbf{{Outdoor Datasets}}} \\
\cmidrule(lr){4-6}\cmidrule(lr){7-9}

& 
& 
& {ScanNet200}
& {ScanNet}
& {ScanNet++}
& {nuScenes}
& {Sem.KITTI}
& {Waymo} \\
\midrule

\rowcolor{gray!16}
& \emph{Supervised}\textsuperscript{\(\dagger\)}
& Volt-B {\scriptsize(94.3\,M)}
& $36.2$
& $77.7$
& $51.7$
& $81.3$
& $70.6$
& $71.5$ \\

\midrule

\multirow{6}{4px}{\rotatebox{90}{\emph{3D only}}}

& LAM3C~{\scriptsize\citep{yamada2026lam3c}}
& PTv3-B {\scriptsize(108.4\,M)}
& $25.3$
& $66.0$
& $34.2$
& --
& --
& -- \\

& MSM~{\scriptsize\citep{hermosilla2025msm}}
& HUNet {\scriptsize(67.6\,M)}
& $26.8$
& $68.7$
& --
& --
& --
& -- \\

& Sonata~{\scriptsize\citep{wu2025sonata}}
& PTv3-B {\scriptsize(108.4\,M)}
& $29.3$
& $72.5$
& \probingsecond{$38.9$}
& $66.1$
& $62.0$
& $60.5$ \\

& PointINS~{\scriptsize\citep{yang2026pointins}}
& PTv3-B {\scriptsize(108.4\,M)}
& $29.6$
& $72.4$
& --
& $74.6$
& $68.0$
& -- \\

& DOS~{\scriptsize\citep{abdelsamad2026dos}}
& PTv3-B {\scriptsize(108.4\,M)}
& \probingsecond{$30.7$}
& \probingsecond{$73.9$}
& --
& \probingsecond{$74.8$}
& \probingsecond{$68.3$}
& \probingsecond{$67.0$} \\

& \textbf{\ours}~{\scriptsize (Ours)}
& Volt-B {\scriptsize(85.7\,M)}
& \probingbest{$35.9$}
& \probingbest{$76.9$}
& \probingbest{$42.4$}
& \probingbest{$76.7$}
& \probingbest{$70.7$}
& \probingbest{$68.6$} \\

\midrule

\multirow{5}{4px}{\rotatebox{90}{\emph{with 2D}}}
& LargeAD~{\scriptsize\citep{kong2025largead}} 
& MinkUNet {\scriptsize(37.9\,M)}
& --
& --
& --
& $47.8$
& $35.8$
& $36.1$ \\

& ScaLR~{\scriptsize\citep{puy2024three}} 
& WaffleIron {\scriptsize(59.4\,M)}
& --
& --
& --
& $67.8$
& $55.8$
& -- \\

& Concerto~{\scriptsize\citep{zhang2025concerto}}
& PTv3-B {\scriptsize(108.4\,M)}
& \probingsecond{$37.4$}
& $77.3$
& \probingsecond{$45.6$}
& $74.2$
& $66.6$
& $62.2$ \\

& Utonia~{\scriptsize\citep{zhang2026utonia}}
& PTv3-B {\scriptsize(137.3\,M)}
& $36.4$
& \probingsecond{$77.7$}
& $44.7$
& \probingsecond{$75.5$}
& \probingsecond{$67.7$}
& \probingsecond{$63.8$} \\

& \textbf{\ours}~{\scriptsize (Ours)}
& Volt-B {\scriptsize(85.7\,M)}
& \probingbest{$40.4$}
& \probingbest{$77.9$}
& \probingbest{$48.2$}
& \probingbest{$80.6$}
& \probingbest{$71.1$}
& \probingbest{$70.6$} \\

\bottomrule
\end{tabularx}
\vspace{-5px}
\caption{
\textbf{Linear Probing 3D Semantic Segmentation (mIoU).}
For each dataset, a linear classifier is trained on the training split with the 3D encoder frozen and evaluated on the validation split.
Methods are grouped by 3D-only pretraining \emph{(top)} and pretraining with additional 2D supervision \emph{(bottom)}.
DOS and PointINS outdoor results use PTv3-S. \textsuperscript{\(\dagger\)} is trained end-to-end with Volt's decoder and serves as a reference.
}
\label{tab:linear_probing_semseg}

\end{table*}
\begin{table*}[t]
\centering
\notfootnotesize
\color{black}

\setlength{\tabcolsep}{3pt}
\renewcommand{\probingbest}[1]{\probingscore{probingbestbg}{\boldmath\textbf{#1}}}
\renewcommand{\arraystretch}{1.08}

\begin{tabularx}{\linewidth}{llYYYYYY}
\toprule
\multirow{2}[2]{*}{\textbf{Method}} & \multirow{2}[2]{*}{\textbf{Backbone}} & \multicolumn{3}{c}{\textbf{Indoor Datasets}}
    & \multicolumn{3}{c}{\textbf{Outdoor Datasets}} \\
\cmidrule(lr){3-5}\cmidrule(lr){6-8}

& 
& ScanNet200
& ScanNet
& ScanNet++
& nuScenes
& Sem.KITTI
& Waymo
\\
\midrule

D-DITR~{\scriptsize\citep{knaebel2025ditr}}
& PTv3-S {\scriptsize(46.2\,M)}
& $37.7$
& $79.2$
& --
& $80.7$
& $69.8$
& $72.1$
\\

Sonata~{\scriptsize\citep{wu2025sonata}}
& PTv3-B {\scriptsize(124.8\,M)}
& $36.8$
& $79.4$
& $49.3$
& $81.7$
& $72.6$
& \probingsecond{$72.9$}
\\

DOS~{\scriptsize\citep{abdelsamad2026dos}}
& PTv3-B {\scriptsize(124.8\,M)}
& $37.1$
& $79.7$
& --
& $81.8$
& \probingsecond{$73.5$}
& \probingbest{$73.9$}
\\

PointINS~{\scriptsize\citep{yang2026pointins}}
& PTv3-B {\scriptsize(124.8\,M)}
& $36.6$
& $79.0$
& --
& $81.1$
& $72.7$
& --
\\

Concerto~{\scriptsize\citep{zhang2025concerto}}
& PTv3-B {\scriptsize(124.8\,M)}
& $39.2$
& $80.7$
& \probingsecond{$50.7$}
& $82.0$
& $71.2$
& $69.2$
\\

Utonia~{\scriptsize\citep{zhang2026utonia}}
& PTv3-B {\scriptsize(157.7\,M)}
& \probingsecond{$39.6$}
& \probingsecond{$81.1$}
& $49.0$
& \probingsecond{$82.2$}
& $72.0$
& $71.4$
\\

\textbf{\ours}~{\scriptsize (Ours)}
& Volt-B {\scriptsize(94.3\,M)} & \probingbest{$42.7$}          & \probingbest{$81.8$}          & \probingbest{$52.6$}          & \probingbest{$83.2$}          & \probingbest{$75.2$}          & $72.9$                        \\

\bottomrule
\end{tabularx}
\vspace{-5px}
\caption{\textbf{Fine-tuning 3D Semantic Segmentation (mIoU).}
Pretrained encoders and task-specific decoders are jointly fine-tuned on the training split and evaluated on the validation split for each dataset.
}
\vspace{-10pt}
\label{tab:finetuning_semseg}
\end{table*}

\subsection{Semantic Segmentation}

\PARbegin{Linear Probing.}
We first evaluate the quality of the representations by freezing the pretrained 3D encoder and training only a linear head for semantic segmentation.
\cref{tab:linear_probing_semseg} (\emph{top}) compares methods using only 3D pretraining.
Here, \ours{} achieves the best linear-probing performance on all six benchmarks.
The gains are particularly pronounced on indoor scenes, improving over the strongest prior result by $+5.2$ mIoU on ScanNet200, $+3.0$ on ScanNet, and $+3.5$ on ScanNet++.
On outdoor scenes, \ours{} outperforms the strongest prior method, DOS, by $+1.9$ mIoU on nuScenes, $+2.4$ on SemanticKITTI, and $+1.6$ on Waymo.
Notably, \ours{} uses an 85.7M-parameter Volt-B encoder, compared with the larger 108.4M-parameter PTv3-B backbone used by most prior methods.

\begin{table*}[t]
\centering
\notfootnotesize
\setlength{\tabcolsep}{1pt}
\renewcommand{\probingbest}[1]{\probingscore{probingbestbg}{\boldmath\textbf{#1}}}
\renewcommand{\arraystretch}{1.08}

\begin{tabularx}{\textwidth}{c>{\hspace{3pt}}l>{\hspace{1pt}}lYYYYYY}
\toprule
 & \multirow{2}[2]{*}{\textbf{Method}} & \multirow{2}[2]{*}{\textbf{Backbone}}
 & \multicolumn{2}{c}{\textbf{ScanNet200}}
 & \multicolumn{2}{c}{\textbf{ScanNet}}
 & \multicolumn{2}{c}{\textbf{ScanNet++}} \\
\cmidrule(lr){4-5}\cmidrule(lr){6-7}\cmidrule(lr){8-9}

&
&
& {mAP$_{50}$} & {mAP}
& {mAP$_{50}$} & {mAP}
& {mAP$_{50}$} & {mAP} \\
\midrule

\multirow{6}{4px}{\rotatebox{90}{\emph{3D only}}}

& LAM3C~{\scriptsize\citep{yamada2026lam3c}}
& PTv3-B {\scriptsize(108.4\,M)}
& --     & $8.3$
& --     & $25.1$
& --     & $11.3$ \\

& MSM~{\scriptsize\citep{hermosilla2025msm}}
& HUNet {\scriptsize(67.6\,M)}
& $8.8$  & --
& $44.4$ & --
& --     & -- \\

& Sonata~{\scriptsize\citep{wu2025sonata}}
& PTv3-B {\scriptsize(108.4\,M)}
& $21.3$ & $10.9$
& $53.9$ & $30.7$
& \probingsecond{$26.5$} & \probingsecond{$14.8$} \\

& DOS~{\scriptsize\citep{abdelsamad2026dos}}
& PTv3-B {\scriptsize(108.4\,M)}
& $20.6$\rlap{$^{\ddagger}$} & $10.9$\rlap{$^{\ddagger}$}
& $54.6$ & $32.5$
& --     & -- \\

& PointINS~{\scriptsize\citep{yang2026pointins}}
& PTv3-B {\scriptsize(108.4\,M)}
& \probingsecond{$24.9$}\rlap{$^{\ddagger}$} & \probingsecond{$13.4$}\rlap{$^{\ddagger}$}
& \probingsecond{$57.6$} & \probingsecond{$34.6$}
& --     & -- \\

& \textbf{\ours}~{\scriptsize (Ours)}
& Volt-B {\scriptsize(85.7\,M)}
& \probingbest{$30.0$} & \probingbest{$15.8$}
& \probingbest{$59.5$} & \probingbest{$35.1$}
& \probingbest{$28.8$} & \probingbest{$16.3$} \\

\midrule

\multirow{3}{4px}{\rotatebox{90}{\emph{with 2D}}}

& Concerto~{\scriptsize\citep{zhang2025concerto}}
& PTv3-B {\scriptsize(108.4\,M)}
& \probingsecond{$27.7$} & \probingsecond{$14.9$}
& \probingsecond{$55.7$} & \probingsecond{$31.1$}
& $31.3$ & \probingbest{$18.2$} \\

& Utonia\textsuperscript{\(\dagger\)}~{\scriptsize\citep{zhang2026utonia}}
& PTv3-B {\scriptsize(137.3\,M)}
& \probingsecond{$27.7$} & $14.2$
& $53.2$ & $29.0$
& \probingsecond{$32.0$} & $17.4$ \\

& \textbf{\ours}~{\scriptsize (Ours)}
& Volt-B {\scriptsize(85.7\,M)}
& \probingbest{$35.0$} & \probingbest{$18.7$}      & \probingbest{$59.7$} & \probingbest{$35.5$}      & \probingbest{$32.7$} & \probingsecond{$18.0$} \\ 

\bottomrule
\end{tabularx}
\vspace{-5px}
\caption{
\textbf{Linear Probing 3D Instance Segmentation.}
PointGroup~\citep{jiang2020pointgroup} is trained with a linear semantic head and a linear offset head on top of the frozen 3D encoder.
\textsuperscript{\(\dagger\)}~Reproduced.
DOS and PointINS results are taken from \citet{yang2026pointins}.
\textsuperscript{\(\ddagger\)}~Single-dataset pretraining with PTv3-S encoder.
}

\label{tab:linear_probing_insseg}

\end{table*}
\begin{table*}[t]
\centering
\notfootnotesize
\color{black}

\setlength{\tabcolsep}{3pt}
\renewcommand{\probingbest}[1]{\probingscore{probingbestbg}{\boldmath\textbf{#1}}}
\renewcommand{\arraystretch}{1.08}

\begin{tabularx}{\linewidth}{llYYYYYY}
\toprule
\multirow{2}[2]{*}{\textbf{Method}} & \multirow{2}[2]{*}{\textbf{Backbone}}
    & \multicolumn{2}{c}{\textbf{ScanNet200}}
    & \multicolumn{2}{c}{\textbf{ScanNet}}
    & \multicolumn{2}{c}{\textbf{ScanNet++}} \\
\cmidrule(lr){3-4}\cmidrule(lr){5-6}\cmidrule(lr){7-8}

&
& mAP$_{50}$ & mAP
& mAP$_{50}$ & mAP
& mAP$_{50}$ & mAP
\\
\midrule

Sonata~{\scriptsize\citep{wu2025sonata}}
& PTv3-B {\scriptsize(124.8\,M)}
& $35.6$ & $25.4$
& $63.9$ & $42.4$
& $36.5$ & $24.6$
\\

DOS$^\ddagger$~{\scriptsize\citep{abdelsamad2026dos}}
& PTv3-S {\scriptsize(46.2\,M)}
& $33.5$ & $22.5$
& $62.0$ & $40.5$
& -- & --
\\

PointINS$^\ddagger$~{\scriptsize\citep{yang2026pointins}}
& PTv3-S {\scriptsize(46.2\,M)}
& $34.8$ & $25.1$
& $63.7$ & $41.5$
& -- & --
\\

Concerto~{\scriptsize\citep{zhang2025concerto}}
& PTv3-B {\scriptsize(124.8\,M)}
& \probingsecond{$38.7$} & \probingbest{$27.4$}
& $64.9$ & \probingsecond{$42.9$}
& \probingsecond{$38.3$} & \probingbest{$26.0$}
\\

Utonia\textsuperscript{\(\dagger\)}~{\scriptsize\citep{zhang2026utonia}}
& PTv3-B {\scriptsize(157.7\,M)}
& $36.1$ & $25.4$
& \probingsecond{$66.4$} & \probingsecond{$42.9$}
& $37.1$ & $25.5$
\\

\textbf{\ours}~{\scriptsize (Ours)}
& Volt-B {\scriptsize(94.3\,M)}
& \probingbest{$41.3$} & \probingsecond{$26.2$}
& \probingbest{$67.0$} & \probingbest{$44.0$}
& \probingbest{$39.5$} & \probingsecond{$25.7$}
\\

\bottomrule
\end{tabularx}
\vspace{-5px}
\caption{\textbf{Fine-tuning 3D Instance Segmentation.}
The pretrained encoder and PointGroup~\citep{jiang2020pointgroup} heads are jointly fine-tuned.
\textsuperscript{\(\dagger\)}~Reproduced.
$^\ddagger$Taken from \citet{yang2026pointins}.
}
\vspace{-10pt}
\label{tab:finetuning_insseg}
\end{table*}

Adding 2D--3D distillation further improves the learned representation (\cref{tab:linear_probing_semseg}, \emph{bottom}).
Compared with our 3D-only model, it improves performance by $+4.5$ mIoU on ScanNet200, $+5.8$ on ScanNet++, $+3.9$ on nuScenes, and $+2.0$ on Waymo.
As a result, \ours{} achieves the best performance across all six benchmarks, outperforming Concerto by $+3.0$ mIoU on ScanNet200 and $+2.6$ on ScanNet++, and Utonia by $+5.1$ on nuScenes and $+6.8$ on Waymo.
With only a linear head on frozen features, \ours{} even surpasses the fully supervised Volt-B with a decoder on ScanNet200, ScanNet, and SemanticKITTI, demonstrating the effectiveness of the design choices in \ours{}.

\PARbegin{Fine-tuning.}
We next evaluate the representations' adaptability to downstream tasks by jointly fine-tuning the pretrained encoder and a lightweight task-specific decoder on each benchmark.
As shown in \cref{tab:finetuning_semseg}, \ours{} achieves the best performance on five of the six benchmarks across indoor and outdoor datasets.
The largest gains are observed on indoor scenes, improving the previous best results by $+3.1$ mIoU on ScanNet200 and $+1.9$ on ScanNet++, while outdoors, \ours{} improves SemanticKITTI by $+1.7$ mIoU.
Notably, these results are obtained with a 94.3M-parameter Volt-B backbone, smaller than the PTv3-B backbones used by most competing methods.
This shows that the representations learned by \ours{} are not only readily usable as frozen features but also provide strong initialization for full downstream fine-tuning.

\subsection{Instance Segmentation}
\label{sec:insseg}
We further evaluate whether the learned representations capture object-level structure beyond per-point semantics.
Following prior work~\citep{wu2025sonata,zhang2025concerto}, we attach PointGroup~\citep{jiang2020pointgroup} to the 3D encoder and evaluate under both linear probing and fine-tuning.
PointGroup predicts a semantic label and an offset vector toward the instance center for each point, and groups points into instances by clustering in the original and offset-shifted coordinate spaces.

\PARbegin{Linear Probing.}
We first freeze the pretrained encoder and train only a linear semantic head and a linear offset head.
As shown in \cref{tab:linear_probing_insseg}, \ours{} with 3D-only pretraining outperforms all prior 3D-only methods on every benchmark, with the largest gain of $+5.1$ mAP$_{50}$ on ScanNet200.
It even surpasses Concerto and Utonia, both of which additionally use 2D supervision, by $+2.3$ mAP$_{50}$ on ScanNet200 and $+3.8$ on ScanNet.
Adding 2D--3D distillation further widens the margin over the strongest 2D-supervised methods to $+7.3$ mAP$_{50}$ on ScanNet200 and $+4.0$ on ScanNet.
On ScanNet++, \ours{} also achieves the best mAP$_{50}$, while trailing Concerto by only 0.2 mAP.

\PARbegin{Fine-tuning.}
We next jointly fine-tune the pretrained encoder and the PointGroup heads.
As shown in \cref{tab:finetuning_insseg}, \ours{} achieves the best mAP$_{50}$ on all three benchmarks, improving over the strongest prior method by $+2.6$ on ScanNet200, $+0.6$ on ScanNet, and $+1.2$ on ScanNet++.
As in linear probing, the gain is largest on ScanNet200, whose long-tailed label set likely benefits most from a strong initialization.
These results are obtained with a 94.3M-parameter Volt-B backbone, smaller than the PTv3-B encoders used by Sonata, Concerto, and Utonia.
Together with the linear-probing results, this confirms that the benefits of \ours{} extend from dense semantic prediction to object-level scene understanding.
To further push performance, we also fine-tune \ours{} with the stronger SPFormer~\citep{sun2023spformer} decoder in \cref{sec:supp_sota_insseg}.

\subsection{Cross-Domain Transfer}
We next evaluate how well the learned representations transfer beyond the settings in which they were pretrained.
We consider two complementary settings: (i) point-cloud registration on datasets unseen during pretraining, which probes geometric correspondence, and (ii) cross-modal semantic probing, in which 2D and 3D representations are each evaluated in the other's native input domain, probing semantic transfer.

\begin{wraptable}[12]{r}{0.45\columnwidth}
\centering
\small
\color{black}
\vspace{-10pt} 

\setlength{\tabcolsep}{3pt}
\renewcommand{\arraystretch}{1.08}

\begin{tabularx}{\linewidth}{lYY}
    \toprule
    \textbf{Method}
    & \textbf{3DMatch}
    & \textbf{3DLoMatch} \\
    \midrule
    FPFH~{\scriptsize\citep{rusu2009fpfh}}
    & \probingsecond{$77.5$}
    & \probingsecond{$36.4$} \\
    \midrule
    Concerto
    & $65.0$
    & $22.7$ \\
    Utonia
    & $76.5$
    & $36.3$ \\
    \textbf{\ours}~{\scriptsize (Ours)}
    & \probingbest{$84.1$}
    & \probingbest{$44.6$} \\
    \bottomrule
\end{tabularx}

\vspace{-5px}
\caption{
\textbf{Point-Cloud Registration.}
Registration success rate (\%) on 3DMatch and 3DLoMatch using frozen encoder features as descriptors with a shared registration backend~\citep{chen2022sc2pcr}.
FPFH is a hand-crafted descriptor.
}
\label{tab:registration}
\end{wraptable}
\PARbegin{Point-Cloud Registration.}
We evaluate whether the learned representations transfer to geometric correspondence on 3DMatch~\citep{zeng20173dmatch} and its low-overlap counterpart, 3DLoMatch~\citep{huang2021predator}, neither of which is part of our pretraining data.
We use the frozen encoder features as point descriptors, form candidate correspondences by nearest-neighbor matching between the two point clouds, and estimate the transformation with SC$^2$-PCR~\citep{chen2022sc2pcr}.
Following~\citet{pan2026rap}, we report the registration success rate.
As shown in \cref{tab:registration}, \ours{} achieves 84.1\% on 3DMatch and 44.6\% on 3DLoMatch, exceeding Utonia by $+7.6$ and $+8.3$ points and Concerto by $+19.1$ and $+21.9$ points, respectively.
\ours{} also outperforms the hand-crafted FPFH descriptor~\citep{rusu2009fpfh} by $+6.6$ and $+8.2$ points, whereas both prior learned representations fall short of it.
The gains are larger in the low-overlap setting, where fewer correspondences are available and descriptor quality matters most.
These results show that the learned representations transfer effectively to geometric correspondence without any registration-specific training.

\begin{wraptable}[15]{r}{0.45\columnwidth}
\centering
\small
\color{black}
\vspace{-10pt} 

\setlength{\tabcolsep}{3pt}
\renewcommand{\arraystretch}{1.08}

\begin{tabularx}{\linewidth}{llYY}
    \toprule
    \textbf{Method}
    & \textbf{Arch.}
    & \textbf{NYUv2}
    & \textbf{ScanNet++} \\
    \midrule
    DINOv2
    & ViT-B/14
    & $56.0$
    & $34.5$ \\
    DINOv2
    & ViT-g/14
    & $59.0$
    & $39.3$ \\
    DINOv3
    & ViT-B/16
    & \probingbest{$62.0$}
    & $34.2$ \\
    \midrule
    Concerto
    & PTv3-B
    & $54.9$
    & \probingsecond{$45.6$} \\
    Utonia
    & PTv3-B
    & $55.9$
    & $44.7$ \\
    \textbf{\ours}~{\scriptsize (Ours)}
    & Volt-B
    & \probingsecond{$59.2$}
    & \probingbest{$48.2$} \\
    \bottomrule
\end{tabularx}

\vspace{-5px}
\caption{
\textbf{Cross-Modal Semantic Probing (mIoU).}
Linear probing with frozen encoders.
On NYUv2, single-view RGB-D images are lifted to 3D for the 3D methods.
On ScanNet++, 2D features are lifted from posed views and aggregated across views for the 2D methods.
}
\label{tab:rgbd_linear_probing}
\end{wraptable}
\PARbegin{Cross-Modal Semantic Probing.}
We further compare 2D and 3D representations by evaluating each in its native domain and in the other's (\cref{tab:rgbd_linear_probing}).
On NYUv2~\citep{silberman2012nyuv2}, we apply the 3D models to single-view RGB-D images by lifting each image into 3D with a monocular geometry estimator~\citep{kong2026moge3}, and linearly probe the frozen features.
Although it is pretrained on full-scene 3D reconstructions, \ours{} reaches 59.2 mIoU, outperforming prior 3D representations by at least $+3.3$ mIoU and even surpassing DINOv2-g, its own 2D teacher, while trailing DINOv3-B by 2.8 mIoU.
Conversely, on ScanNet++~\citep{yeshwanthliu2023scannetpp}, we apply the 2D VFMs to 3D by lifting their features from posed RGB views and aggregating them across views before linear probing.
Here, \ours{} reaches 48.2 mIoU, outperforming DINOv2-g by $+8.9$, DINOv3-B by $+14.0$, and the strongest prior 3D representation, Concerto, by $+2.6$.
Together, these results show that \ours{} is competitive with 2D VFMs even in their native single-view setting, and clearly surpasses them in multi-view, full-scene 3D scene understanding.

\subsection{Analyses}

\begin{wraptable}[10]{r}{0.43\columnwidth}
\centering
\small
\vspace{-10pt} 

\begin{tabularx}{\linewidth}{
    >{\raggedright\arraybackslash}X
    >{\centering\arraybackslash}p{0.25\linewidth}
}
\toprule
\textbf{Predictor Design}
& \textbf{ScanNet200} \\
\midrule

\defaultrow
Cross-attention only decoder & $\mathbf{35.9}$ \\
Query-to-query attention     & $34.2$ \\
Full self-attention decoder  & $33.0$ \\
Registers as context         & $35.3$ \\

\bottomrule
\end{tabularx}
\vspace{-5px}
\caption{
\textbf{Masked Predictor.}
Semantic segmentation linear-probing performance (mIoU) for different predictor designs.
}
\label{tab:predictor_ablation}
\end{wraptable}
\paragraph{Predictor Analysis.}
We analyze the masked predictor design in \cref{tab:predictor_ablation}.
Our cross-attention-only decoder performs best at 35.9 mIoU.
Adding query-to-query attention reduces performance to 34.2, while a full self-attention decoder causes a further drop to 33.0.
Since each query carries the position of its masked region, any interaction among queries exposes the geometry of the masked region and can reintroduce the geometric shortcut we try to avoid.
Finally, we find that allowing decoder queries to additionally attend to register tokens, rather than only the 3D patch tokens, reduces performance to 35.3 mIoU.

\begin{wraptable}[7]{r}{0.43\columnwidth}
\centering
\small
\setlength{\tabcolsep}{4pt}
\vspace{-10pt}

\begin{tabularx}{\linewidth}{
    >{\centering\arraybackslash}c
    >{\centering\arraybackslash}X
    @{}p{10pt}@{}
    >{\centering\arraybackslash}c
    >{\centering\arraybackslash}X
}
\noalign{\vskip-\aboverulesep}

\cmidrule[\heavyrulewidth](r{-\tabcolsep}){1-2}
\cmidrule[\heavyrulewidth](l{-\tabcolsep}){4-5}

\textbf{\shortstack{Block\\Size}}
& \raisebox{+0.5\baselineskip}{\textbf{ScanNet200}}
& &
\textbf{\shortstack{Mask\\Ratio}}
& \raisebox{+0.5\baselineskip}{\textbf{ScanNet200}} \\

\cmidrule[\lightrulewidth](r{-\tabcolsep}){1-2}
\cmidrule[\lightrulewidth](l{-\tabcolsep}){4-5}

$60$\,cm
& $35.7$
& &
$45\%$
& $35.2$ \\

\cellcolor{defaultgray}$80$\,cm
& \cellcolor{defaultgray}$\mathbf{35.9}$
& 
& \cellcolor{defaultgray}$55\%$
& \cellcolor{defaultgray}$\mathbf{35.9}$ \\

$100$\,cm
& $34.5$
& &
$65\%$
& $34.8$ \\

\cmidrule[\heavyrulewidth](r{-\tabcolsep}){1-2}
\cmidrule[\heavyrulewidth](l{-\tabcolsep}){4-5}

\end{tabularx}

\vspace{-5px}
\caption{
\textbf{Masking.}
Linear probing performance across block sizes and mask ratios.
}
\label{tab:mask_ablation}
\vspace{-10pt}
\end{wraptable}
\PARbegin{Masking Analysis.}
We analyze the masking configuration in \cref{tab:mask_ablation}.
A block size of 80\,cm performs best at 35.9 mIoU, while 55\% masking outperforms both lower and higher ratios.
This suggests that moderate spatial masking provides a good balance between prediction difficulty and visible context.
We therefore use 80\,cm blocks with a 55\% masking ratio throughout.

\section{Conclusion}
We introduce \ours{}, a 3D representation learning framework that revisits the key design choices of current approaches.
Building on the Volume Transformer, \ours{} constructs reliable targets using clean global views and a target head decoupled from the student.
For masked modeling, it removes masked tokens from the encoder, preventing the student from exploiting their locations and requiring it to infer their representations from the remaining scene context through a separate predictor.
Finally, it distills knowledge from a 2D VFM by learning view-consistent prototype assignments and using them to supervise the corresponding 3D representations.
Together, these changes substantially improve frozen representations, provide a strong initialization for fine-tuning, and enable transfer to unseen datasets.
We hope this work helps bring 3D representation learning closer to the powerful general-purpose representations that scalable self-supervised pretraining has unlocked in 2D.

\subsubsection*{Acknowledgments}
We acknowledge funding by BMFTR project \enquote{WestAI} (grant no.\ \texttt{16IS22094D}) and the EU project \enquote{JUPITER AI Factory} (grant no.\ \texttt{101250682}).
Computations were performed using resources granted by RWTH Aachen under projects \texttt{thes2360} and \texttt{rwth1968} and by the Gauss Centre for Supercomputing e.V. through the John von Neumann Institute for Computing on the GCS Supercomputer JUWELS at the Jülich Supercomputing Centre.

\bibliography{ARXIV/iclr2027_conference}
\bibliographystyle{ARXIV/iclr2027_conference}

\appendix
\newpage
\section{Supplementary}

\subsection{Extended Related Work}

\PARbegin{2D Self-supervised Learning.}
Self-supervised representation learning in 2D vision has evolved from contrastive objectives toward teacher--student self-distillation and masked prediction.
Early contrastive methods such as MoCo~\citep{he2020moco} and SimCLR~\citep{chen2020simclr} learn invariances by pulling together representations of different augmentations of the same image while pushing apart those of different images.
SwAV~\citep{caron2020swav} replaces pairwise comparisons with online clustering and swapped prediction, matching cluster assignments across augmented views, and introduces the multi-crop strategy.
BYOL~\citep{grill2020byol} shows that negative samples are not required, training an online network to predict the representations of a slowly updated target network.
Building on these ideas, DINO~\citep{caron2021dino} applies EMA-based teacher--student self-distillation to Vision Transformers~\citep{dosovitskiy2021vit}, matching teacher and student prototype distributions across multiple image crops and showing strong semantic structure in the learned features.

In parallel, masked image modeling emerged as an effective pretraining paradigm.
BEiT~\citep{bao2021beit} replaces masked patches with a learned mask embedding and trains the model to predict discrete visual tokens produced by a separate tokenizer.
MAE~\citep{he2022mae} instead reconstructs masked patches in pixel space. Its encoder processes only the visible patches, and a lightweight decoder recovers the masked content from the visible context.
CrossMAE~\citep{fu2024crossmae} further shows that this decoder does not require self-attention among masked tokens and replaces it with cross-attention to the visible tokens.
data2vec~\citep{baevski2022data2vec} and I-JEPA~\citep{assran2023ijepa} move the prediction target to latent space, regressing EMA teacher representations at masked locations.
iBOT~\citep{zhou2021ibot} brings masked prediction into the self-distillation framework by predicting teacher prototype assignments at masked patch locations, extending DINO-style supervision from image-level to dense patch-level targets.
DINOv2~\citep{oquab2023dinov2} and DINOv3~\citep{simeoni2025dinov3} scale this family of objectives and demonstrate that self-supervised ViTs can serve as strong general-purpose visual representations across a broad range of downstream tasks.
More recently, CAPI~\citep{darcet2025capi} revisits the construction of masked prediction targets by clustering latent patch representations and decoupling target construction from prediction.
Our work builds on this line of research and revisits how the core design choices of self-distillation should be formulated for sparse 3D scenes: the encoder, local prediction, masked prediction, and the construction of prediction targets.

\PARbegin{3D Object-level Self-supervised Learning.}
For object-level point clouds, most self-supervised methods adapt masked image modeling to 3D.
Point-BERT~\citep{yu2022pointbert} predicts discrete tokens of masked point patches produced by a pretrained tokenizer.
Point-MAE~\citep{pang2022pointmae} instead reconstructs masked patches directly and moves the mask tokens from the encoder to the decoder to avoid leaking the positions of masked patches.
Point2Vec~\citep{knaebel2023point2vec} adapts data2vec~\citep{baevski2022data2vec} to point clouds and similarly predicts teacher representations of masked patches with a separate predictor rather than in the student encoder.
AsymDSD~\citep{leijenaar2025asymdsd} combines latent masked prediction with invariance learning, adapts SwAV's~\citep{caron2020swav} multi-crop strategy to object-level point clouds, and disables self-attention among mask queries in its predictor.
Its predictor design is the closest to ours.
These works establish that the positions of masked regions can act as a shortcut for masked modeling on object-level point clouds.
Scene-level methods, however, keep mask tokens in the encoder~\citep{wu2025sonata,zhang2025concerto}, although in sparse scenes, voxel occupancy directly encodes the geometry of the masked region and thus exposes this shortcut.
We revisit this design choice at the scene level, remove masked-region geometry from both the encoder and the predictor, and augment masked prediction with the decoupled prototype targets shared with local prediction.

\subsection{Implementation Details}
\label{app:implementation}
\PARbegin{Input and View Construction.}
For the first stage, we use the real-scene subset of Sonata's pretraining data~\citep{wu2025sonata}, i.e., excluding its synthetic datasets.
Indoor point clouds are voxelized at 2\,cm and represented by RGB color and surface normals.
Outdoor point clouds are voxelized at 5\,cm and represented by LiDAR intensity and normals estimated from local neighborhoods.
From each scene, we sample two global crops with ratios of 0.4--1.0 and four local crops with ratios of 0.1--0.4, capping indoor crops at 65{,}536 voxels and outdoor crops at 6{,}000 tokens.
All views independently receive standard geometric augmentations including scaling, rotations, axis flips, and coordinate jitter.
Strong corruptions are applied only to the student's local views.
For indoor scenes, we randomly select three of elastic distortion, point dropout, color jitter, and chromatic translation.
For outdoor scenes, we randomly select two of normal dropout, point dropout, elastic distortion, and 1\,m block dropout.
We partition each student's global view into cubic blocks with a side length of eight tokens (80\,cm indoors and 2\,m outdoors) and mask 55\% of the blocks.
Cross-view token correspondences are established by nearest-neighbor matching of token centers in the original coordinate frame, rejecting matches farther apart than 2 tokens.

\PARbegin{Architecture.}
The Volt-B encoder uses 16 register tokens, LayerScale initialized to $10^{-5}$, and a stochastic-depth rate of 0.1.
The masked predictor consists of four cross-attention layers of width 768 with 12 heads.
For each global view, up to 512 queries are sampled at random from the masked tokens.
The student's local prediction head is a $768 \to 2048 \to 128$ MLP with GELU activations, followed by $\ell_2$ normalization and a bias-free linear layer with 4096 outputs whose weights are initialized from a Gaussian with unit standard deviation.
The masked prediction head is a separate bias-free linear layer that maps the $\ell_2$-normalized predictor output directly to the 4096 prototype logits, without an MLP.
We apply no temperature to the student logits.
The teacher's target head consists of 4096 learned prototypes, with teacher temperature $\tau_t = 0.05$, clustering prediction temperature $\tau_C = 0.1$, and three Sinkhorn--Knopp iterations.

\PARbegin{Self-Supervised Objective.}
All terms are cross-entropy losses against detached targets.
The masked loss comprises a within-view and a cross-view term, where the former uses the teacher assignments of the same global view and the latter those of the corresponding tokens in the other global view.
The overall objective is
\begin{equation}
  \mathcal{L}_{\text{SSL}} = 0.25\,\mathcal{L}_{\text{local}} + 0.25\,\mathcal{L}_{\text{masked}} + 0.25\,\mathcal{L}_{\text{sc}}.
\end{equation}
As described in \cref{sec:local-prediction}, the swapped clustering loss $\mathcal{L}_{\text{sc}}$ updates only the teacher prototypes and does not backpropagate into either encoder.

\PARbegin{Stage 1 Optimization.}
We train for 200k steps with AdamW, a batch size of 32, a learning rate of $8 \times 10^{-4}$ after 10k linear warmup steps, a weight decay of 0.05, and gradient clipping at 3.0.
The teacher prototypes use half the learning rate.
The teacher EMA momentum is fixed at 0.997.
Indoor training takes 24 hours on 8 H100 GPUs, and outdoor training takes 24 hours on 16 H100 GPUs.

\PARbegin{Image Prototypes.}
Before the second stage, we learn the image prototypes on the frozen DINOv2-g/14~\citep{oquab2023dinov2} features of the stage-2 training data, using the swapped clustering loss between image patches that observe a shared valid 3D point.
We use 4096 prototypes and a teacher temperature of 0.04, and train for 30k steps.
Because a single batch contains too few patches for well-balanced Sinkhorn--Knopp assignments, we compute them over a queue of up to 16{,}384 image features per GPU.
The image prototypes are then frozen.

\PARbegin{Stage 2 Data and Augmentations.}
Following Concerto~\citep{zhang2025concerto}, each indoor sample pairs a point cloud with up to four posed RGB images at $518 \times 518$ resolution, drawn uniformly from ScanNet, ScanNet++, ARKitScenes, and RealEstate10K.
For outdoor scenes, we use all six nuScenes cameras, all five Waymo cameras, and the left SemanticKITTI camera, resizing images to a width of 672 pixels while preserving the aspect ratio and padding them to a multiple of 14.
In both settings, the second stage uses a single local crop with a ratio of 0.3--0.7.
Crops in both settings are capped at 8{,}192 tokens.
Indoor local views additionally receive color dropout and unrestricted $\mathrm{SO}(3)$ rotations, each with probability 0.3.

\PARbegin{Stage 2: 2D--3D Correspondences and Loss.}
We obtain 2D--3D correspondences by projecting 3D points into the posed images.
Indoors, we additionally require agreement with the image's depth map within 2\,cm.
In both settings, we discard points more than 5\,cm behind the point closest to the camera among those projected into the same image patch, which removes occluded points.
Following Concerto's patch-level alignment, we average the student's 3D features over all points that project into the same image patch and pass the result through a dedicated prediction head to obtain logits over the image prototypes.
These are trained with cross-entropy against the Sinkhorn--Knopp assignments of the corresponding DINOv2 patch features.
The 3D features come from a separate student forward pass on the unmasked first global view.
The 2D--3D loss has weight 1.0 and is added to $\mathcal{L}_{\text{SSL}}$.

\PARbegin{Stage 2 Optimization.}
We initialize from the first-stage checkpoint and train for 30k steps, decaying the learning rate from $8 \times 10^{-4}$ to $8 \times 10^{-7}$ with a cosine schedule.
Indoor training takes roughly 9 hours on 8 H100 GPUs, and outdoor training takes roughly 8 hours on 16 H100 GPUs.

\PARbegin{Linear Probing.}
For encoders with a PTv3 backbone, we upsample the features from all encoder stages to the finest resolution and concatenate them for per-point features.
For \ours{}, we concatenate the features from the last three encoder layers for each token and assign the resulting token feature to each point in its patch.

\subsection{Instance Segmentation with a State-of-the-Art Decoder}
\label{sec:supp_sota_insseg}
The instance segmentation results in Sec.~\ref{sec:insseg} follow prior 3D representation learning work~\citep{wu2025sonata,zhang2025concerto} and use PointGroup~\citep{jiang2020pointgroup} heads, which lag behind recent instance segmentation methods.
To assess whether the benefits of VINO persist with a modern decoder, we fine-tune it with SPFormer~\citep{sun2023spformer} using three different initializations of the encoder in Tab.~\ref{tab:sota_insseg}.
The first is a Volt-B checkpoint pretrained for semantic segmentation on ScanNet200, ScanNet++, and ARKitScenes (\emph{Supervised}).
The second is our first-stage, 3D-only checkpoint (\emph{VINO-3D}), and the third is our final checkpoint after 2D--3D distillation (\emph{VINO}).
In all cases, the encoder and decoder are fine-tuned jointly with the same schedule and hyperparameters.

\begin{wraptable}{r}{0.5\linewidth}
\vspace{-12pt}
\centering
\notfootnotesize
\color{black}

\setlength{\tabcolsep}{3pt}
\renewcommand{\probingbest}[1]{\probingscore{probingbestbg}{\boldmath\textbf{#1}}}
\renewcommand{\arraystretch}{1.08}

\begin{tabularx}{\linewidth}{lYYYYYY}
\toprule
\multirow{2}[2]{*}{\textbf{Initialization}}
    & \multicolumn{2}{c}{\textbf{ScanNet200}}
    & \multicolumn{2}{c}{\textbf{ScanNet}}
    & \multicolumn{2}{c}{\textbf{ScanNet++}} \\
\cmidrule(lr){2-3}\cmidrule(lr){4-5}\cmidrule(lr){6-7}

& mAP$_{50}$ & mAP
& mAP$_{50}$ & mAP
& mAP$_{50}$ & mAP
\\
\midrule

Supervised
& 48.4 & 38.6
& 78.3 & 59.2
& 53.1 & 34.5
\\

\ours{}-3D
& 50.3 & 39.2
& 82.9 & 64.4
& 54.4 & 35.8
\\

\textbf{\ours}
& 55.7 & 43.3
& 83.4 & 64.7
& 55.8 & 37.0
\\

\bottomrule
\end{tabularx}
\vspace{-5px}
\caption{\textbf{Instance Segmentation with SPFormer.}
Fine-tuning with different encoder initializations.}
\label{tab:sota_insseg}
\vspace{-10pt}
\end{wraptable}
As shown in Tab.~\ref{tab:sota_insseg}, both VINO checkpoints outperform supervised pretraining on all three benchmarks.
With 3D-only pretraining, VINO-3D improves over the supervised initialization by $+1.9$ mAP$_{50}$ on ScanNet200, $+4.6$ on ScanNet, and $+1.3$ on ScanNet++.
Adding 2D--3D distillation improves performance further on every benchmark, raising the gains to $+7.3$ mAP$_{50}$ on ScanNet200, $+5.1$ on ScanNet, and $+2.7$ on ScanNet++.
Together, these results show that the improvements from VINO are not tied to a particular decoder: they carry over to a state-of-the-art instance segmentation architecture and exceed those of fully supervised pretraining.

\subsection{Ablations on ScanNet++}
\begin{wraptable}[21]{r}{0.5\textwidth}
\vspace{-10pt}
\small
\centering
\setlength{\tabcolsep}{5pt}
\renewcommand{\arraystretch}{1.08}

\begin{tabular*}{\linewidth}{
    @{\extracolsep{\fill}}
    lcc
    @{}
}
\toprule
& \textbf{ScanNet200} & \textbf{ScanNet++} \\
\midrule

Sonata\textsuperscript{\(\dagger\)} (PTv3) & $28.3$ & $38.0$ \\
Sonata (Volt)                              & $29.6$ & $39.1$ \\
Simpler scheduling                         & $29.5$ & $39.1$ \\

\midrule
No masked loss                             & $28.9$ & $38.4$ \\
Asymmetric augm.                           & $31.3$ & $39.3$ \\
Decoupled head                             & $32.8$ & $40.9$ \\

\midrule
Local pred.                                & $32.8$ & $40.9$ \\
+ Masked pred.                             & $35.6$ & $42.2$   \\
+ Fix artifacts                            & $35.9$ & $42.4$ \\

\midrule
Align from start                           & $37.2$ & $44.8$ \\
Raw embeddings                             & $38.6$ & $46.1$ \\
Prototypes                                 & $40.4$ & $48.2$ \\

\bottomrule
\end{tabular*}

\caption{\small\textbf{Additional Ablations.}
Semantic segmentation linear-probing performance (mIoU) on ScanNet200 and ScanNet++.
\textsuperscript{\(\dagger\)} is reproduced.
ScanNet200 results correspond to
Tabs.~\ref{tab:baseline}, \ref{tab:local}, \ref{tab:masked}, and \ref{tab:2d3ddistill}.}
\label{tab:supp_ablations}
\end{wraptable}
We repeat the main-paper ablations on ScanNet++~\citep{yeshwanthliu2023scannetpp} to test whether the gains transfer beyond ScanNet200~\citep{rozenberszki2022scannet200}.
As shown in Tab.~\ref{tab:supp_ablations}, every design choice that improves ScanNet200 also improves ScanNet++.
Replacing PTv3~\citep{wu2024ptv3} with Volt~\citep{yilmaz2026volume} improves the Sonata~\citep{wu2025sonata} baseline by $+1.1$ mIoU, and simplifying the schedule has no effect.
For local prediction, asymmetric augmentations and the decoupled head add $+0.9$ and $+1.6$ mIoU; the first gain is smaller than on ScanNet200, while the second is comparable.
For masked prediction, the masked objective and voxel averaging add $+1.3$ and $+0.2$ mIoU, following the same trend as on ScanNet200.
For 2D--3D distillation, two-stage training outperforms aligning from the start by $1.3$ mIoU, and view-consistent prototypes outperform direct feature alignment by $2.1$ mIoU, slightly more than on ScanNet200.
Overall, these results indicate that the benefits of our design choices are not specific to a single benchmark.

\begin{figure*}[t]
    \vspace{-20pt}
    \centering
\setlength{\tabcolsep}{2pt}
    \begin{tabular}{ccccc}
        \small\textbf{Input Image} &
        \small\textbf{DINOv3 Base} &
        \small\textbf{Concerto} &
        \small\textbf{Utonia} &
        \small\textbf{Ours} \\[3pt]
        \includegraphics[trim=7bp 7bp 7bp 7bp,clip,width=0.19\textwidth]{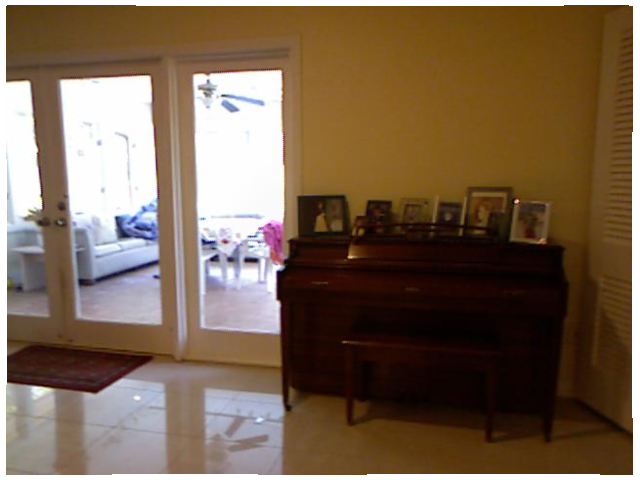} &
        \includegraphics[trim=7bp 7bp 7bp 7bp,clip,width=0.19\textwidth]{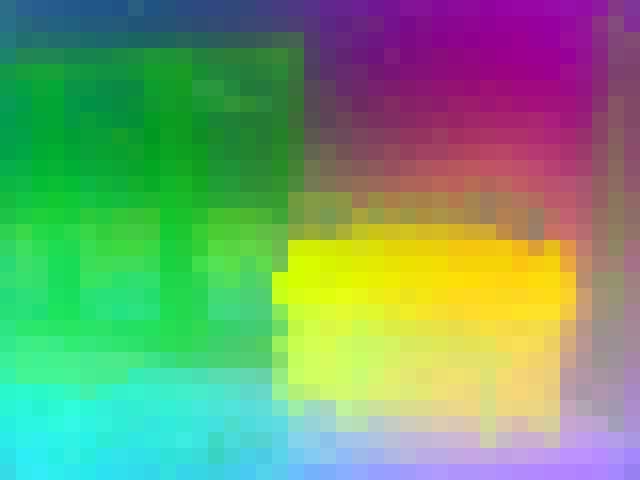} &
        \includegraphics[trim=7bp 7bp 7bp 7bp,clip,width=0.19\textwidth]{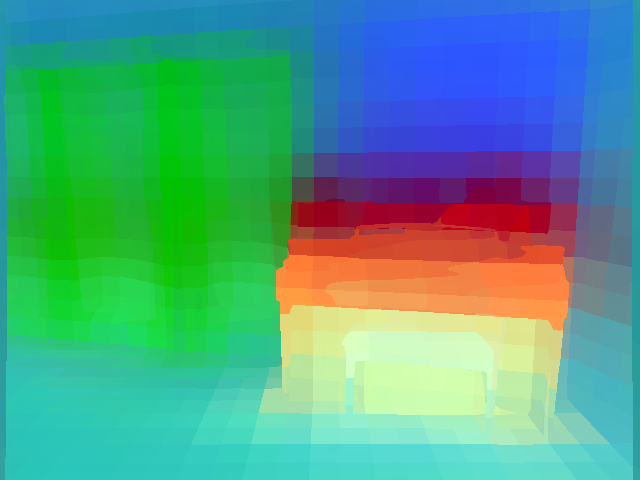} &
        \includegraphics[trim=7bp 7bp 7bp 7bp,clip,width=0.19\textwidth]{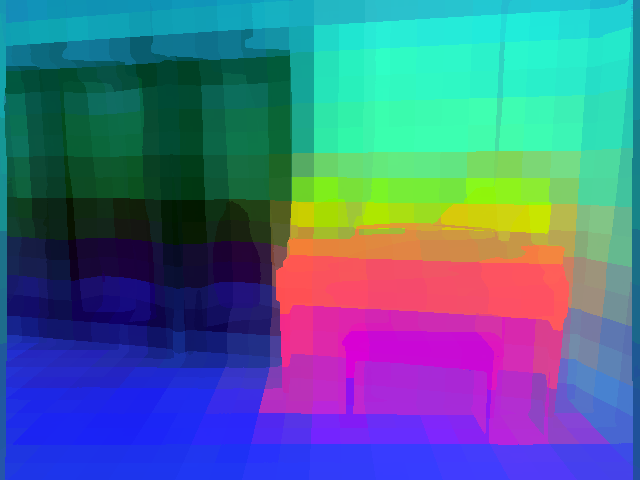} &
        \includegraphics[trim=7bp 7bp 7bp 7bp,clip,width=0.19\textwidth]{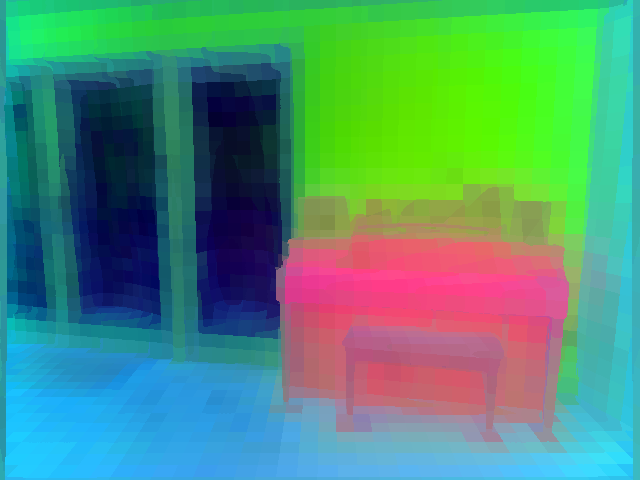}
    \end{tabular}
    \caption{
\textbf{Qualitative Transfer to NYUv2.} PCA of frozen features for a single NYUv2 image. For the 3D models, the image is lifted to 3D.
    }
    \label{fig:nyu_quali}
\end{figure*}

\subsection{Point Cloud Registration With PTv3}
In \cref{tab:registration}, we compare point cloud encoders using a shared registration backend.
Utonia and Concerto use hierarchical PTv3 encoders.
Similar to what they do for linear probing, we upsample features from all encoder stages to the finest resolution and concatenate them.
We then $\ell_2$-normalize the resulting per-point descriptors for registration.
This leads to poor registration performance in our experiments, since deeper stages with larger feature norms contribute more to descriptor similarity.
We therefore $\ell_2$-normalize each stage's features before concatenation, followed by normalization of the concatenated descriptor.
This increases registration success rate on 3DMatch/3DLoMatch from 62.0/18.5\% to 76.5/36.3\% for Utonia and from 54.4/13.4\% to 65.0/22.7\% for Concerto.
We report these stage-normalized results in Tab.~\ref{tab:registration}.

\subsection{Representation Artifacts}
\begin{wrapfigure}[12]{r}{0.5\linewidth}
    \vspace{-28pt}
    \centering
    \definecolor{histbar}{HTML}{4565E3}
    \begin{tikzpicture}
        \begin{groupplot}[
            group style={
                group size=3 by 1,
                horizontal sep=10pt,
                y descriptions at=edge left,
            },
            scale only axis,
            width=0.255\linewidth,
            height=0.25\linewidth,
            xmin=0, xmax=1000,
            ymode=log, ymin=1e-8, ymax=1,
            log origin=infty,
            xtick={0,400,800},
            xticklabels={
                \textnormal{0},
                \textnormal{400},
                \textnormal{800}
            },
            ytick={1e-8,1e-7,1e-6,1e-5,1e-4,1e-3,1e-2,1e-1,1e0},
            yticklabels={
                \textnormal{10\textsuperscript{-8}},,,,
                \textnormal{10\textsuperscript{-4}},,,,
                \textnormal{10\textsuperscript{0}}
            },
            minor y tick num=8,
            ytick pos=left,
            xtick pos=bottom,
            tick align=outside,
            major tick length=3pt,
            minor tick length=1.8pt,
            xticklabel style={font=\tiny},
            yticklabel style={font=\tiny},
            title style={font=\scriptsize, yshift=-2pt},
            xlabel style={font=\footnotesize, yshift=2pt},
            every axis plot/.append style={fill=histbar, draw=none},
        ]
            \nextgroupplot[title={no reg.\ + no avg}]
            \addplot[ybar interval]
                table[
                    x=bin_left,
                    y=no_registers_no_avg_fraction,
                    col sep=comma
                ] {ARXIV/figures/data/token_norm_histogram.csv};

            \nextgroupplot[
                title={16 reg.\ + no avg},
                xlabel={$\ell_2$ norm},
                ytick=\empty,
                minor y tick num=0
            ]
            \addplot[ybar interval]
                table[
                    x=bin_left,
                    y=registers_no_avg_fraction,
                    col sep=comma
                ] {ARXIV/figures/data/token_norm_histogram.csv};

            \nextgroupplot[
                title={16 reg.\ + avg},
                ytick=\empty,
                minor y tick num=0
            ]
            \addplot[ybar interval]
                table[
                    x=bin_left,
                    y=registers_avg_fraction,
                    col sep=comma
                ] {ARXIV/figures/data/token_norm_histogram.csv};
        \end{groupplot}
    \end{tikzpicture}
    \caption{
        \textbf{Token Norm Histograms.}
        Distribution of token $\ell_2$ norms, with frequency shown on a logarithmic scale.
        Registers remove many high-norm outliers; voxel averaging removes most of the remaining tail.
    }
    \label{fig:norm_histograms}
    \vspace{-6pt}
\end{wrapfigure}
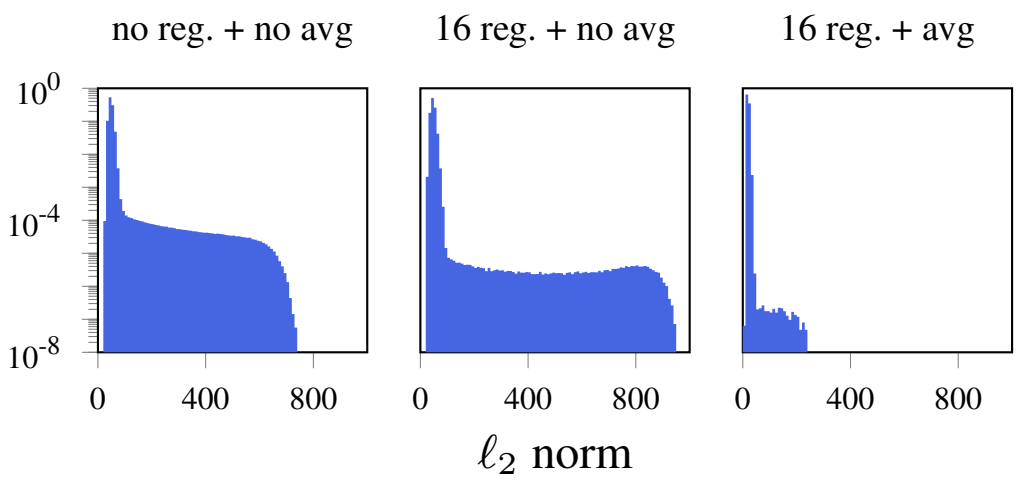
Fig.~\ref{fig:norm_histograms} shows the distribution of token $\ell_2$ norms under our artifact-mitigation steps.
Without registers or voxel averaging, the representation contains a pronounced tail of high-norm tokens.
Adding register tokens removes many of these outliers, while averaging voxel features within each token suppresses most of the remaining tail.
This supports our observation that representation artifacts arise both from isolated high-norm tokens and from feature magnitude scaling with voxel occupancy.

\begin{figure}[p]
    \centering
    \includegraphics[width=\linewidth]{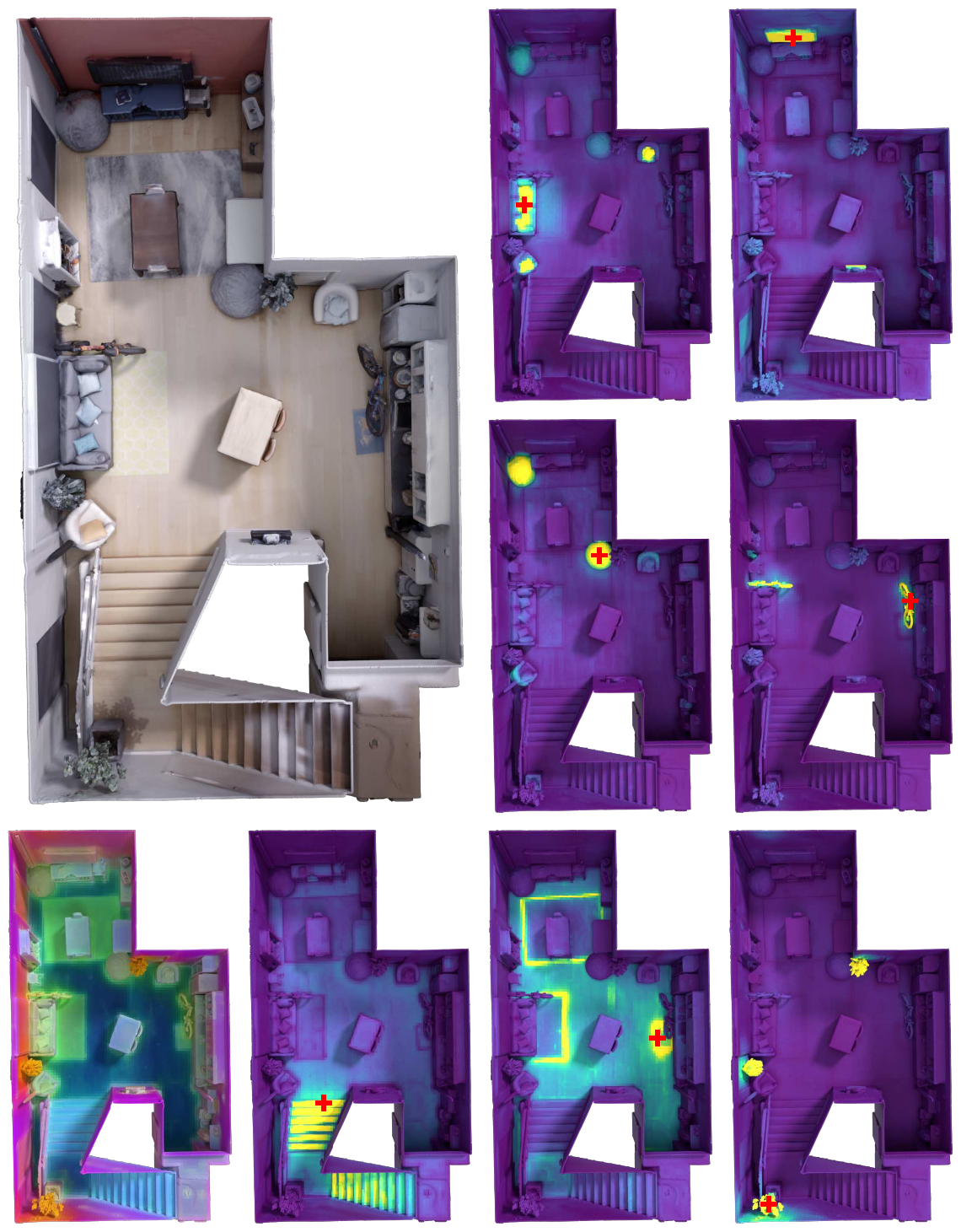}
    \caption{
        \textbf{Qualitative results on Replica.} \ours{} features of a large indoor scene from Replica~\citep{straub2019replica}. We visualize PCA on the bottom left and cosine similarity maps between tokens marked with a red cross and the other 3D tokens.
    }
    \label{fig:replica_quali}
\end{figure}

\begin{figure}[p]
    \centering
    \includegraphics[width=\linewidth]{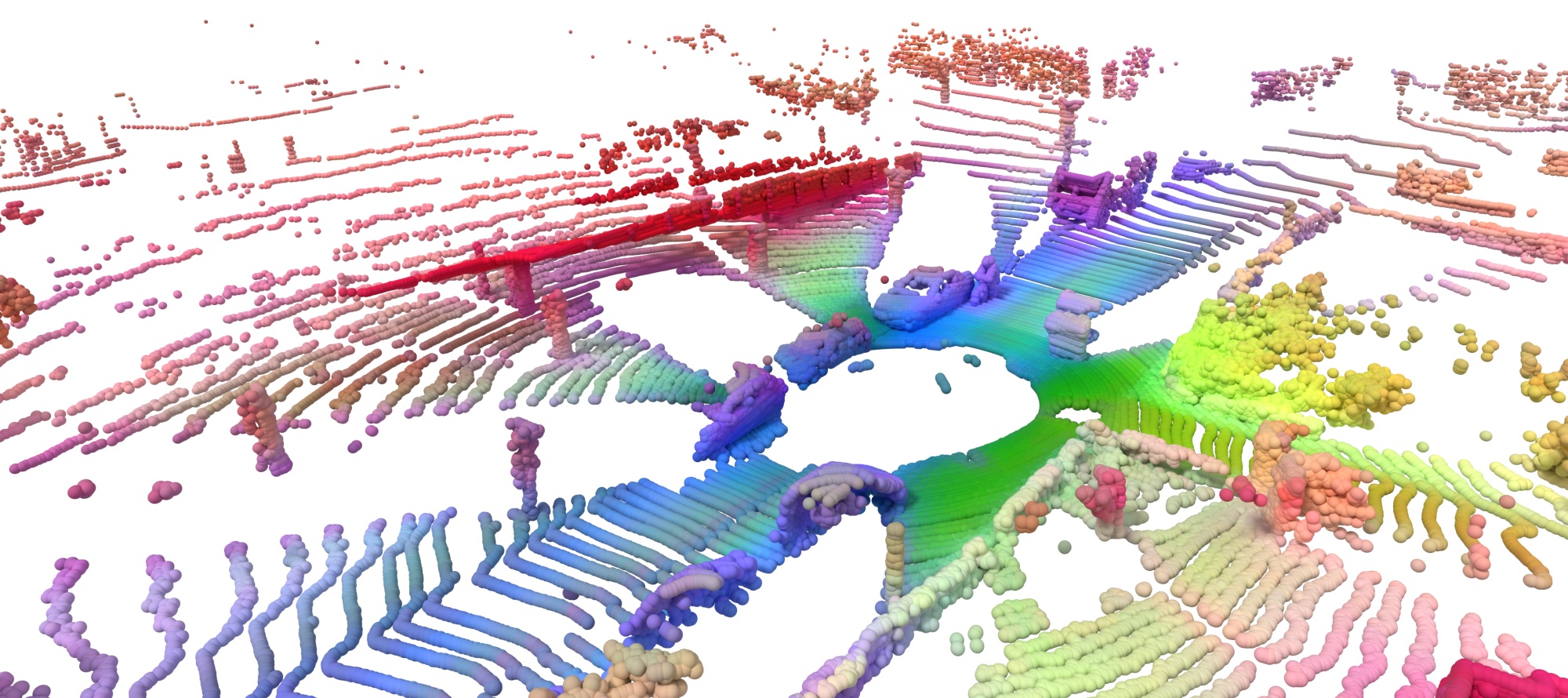}\\[4pt]
    \includegraphics[width=\linewidth]{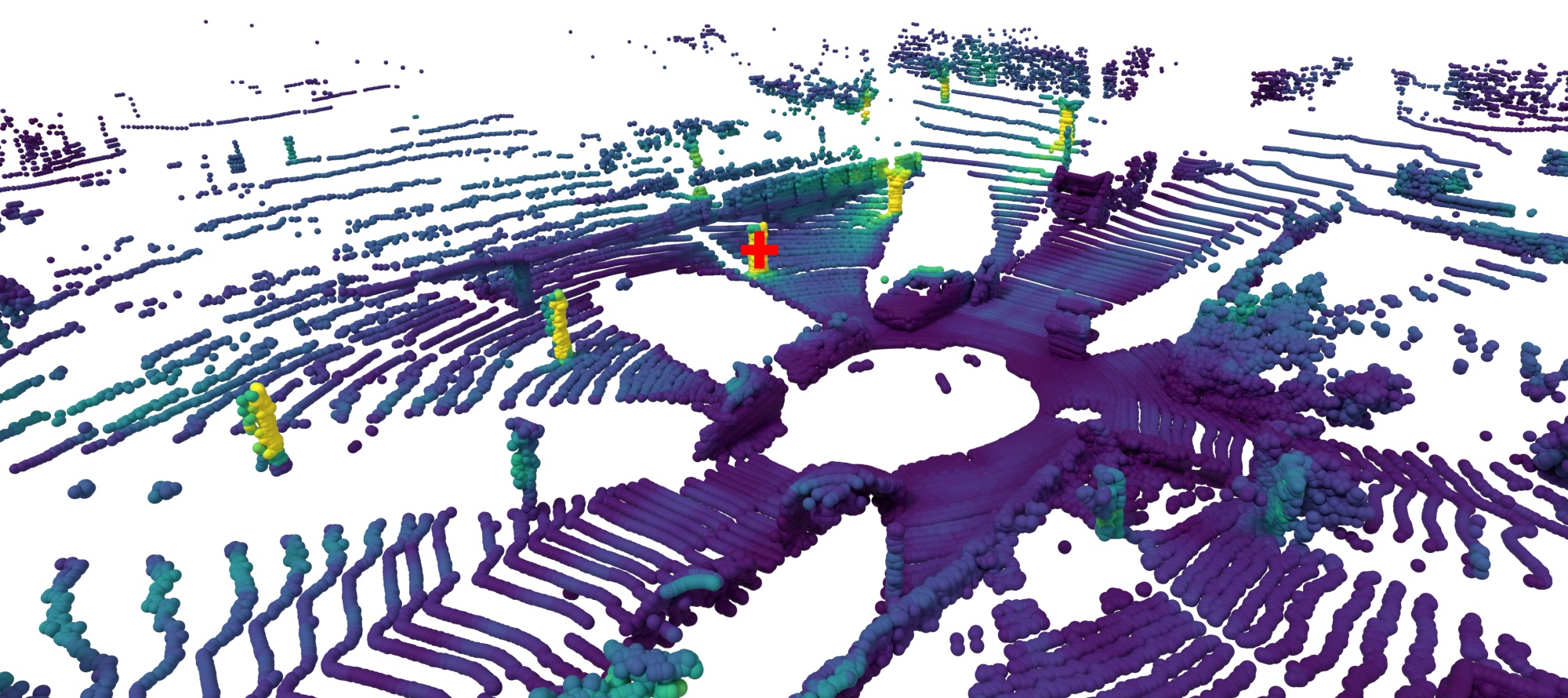}\\[4pt]
    \includegraphics[width=\linewidth]{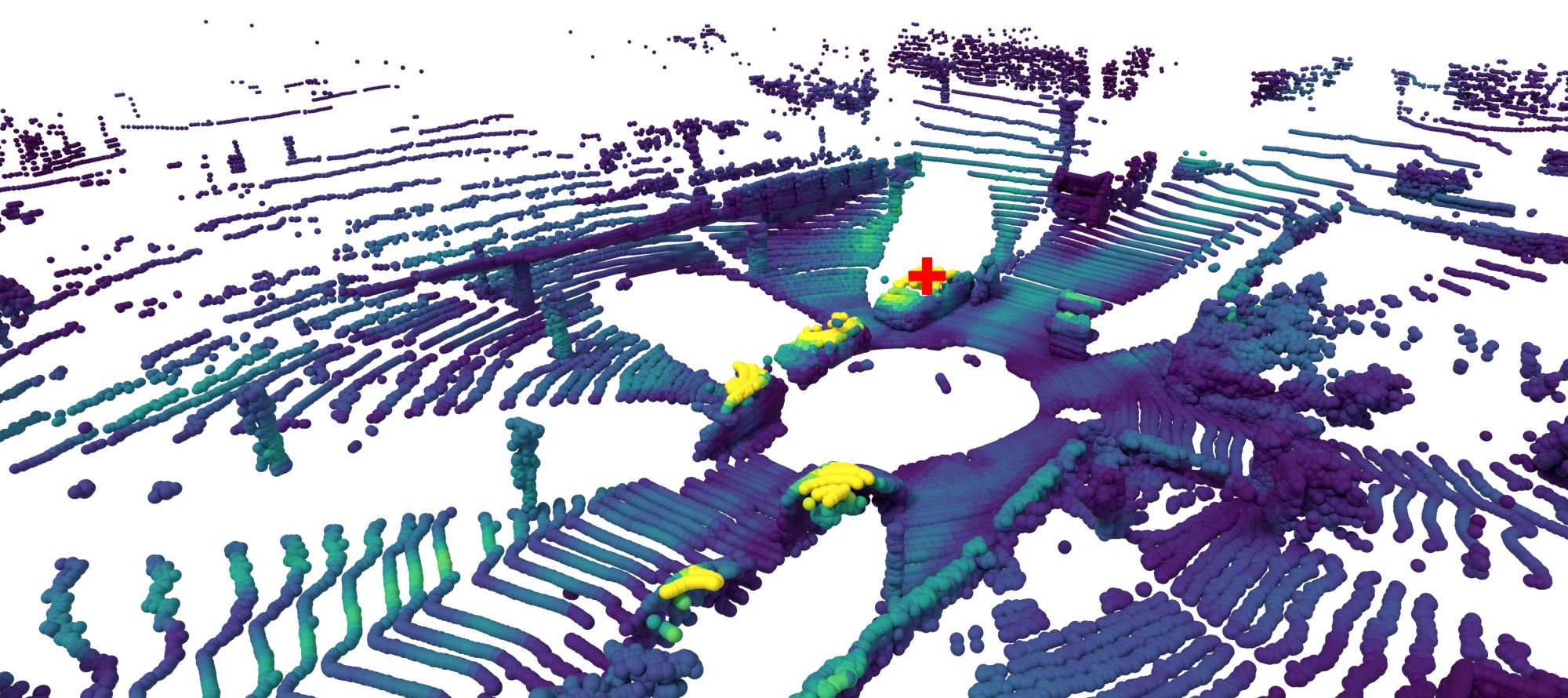}
    \caption{
        \textbf{Qualitative results on SemanticKITTI.} VINO features of an outdoor scene from SemanticKITTI~\citep{geiger2012kitti,behley2019semantickitti}. We visualize PCA (top) and cosine similarity maps between the token marked with a red cross and all other 3D tokens (middle and bottom).
    }
    \label{fig:kitti_quali}
\end{figure}

\subsection{Qualitative Results}
\label{sec:supp_qualitative}

\PARbegin{Transfer to Single-View RGB-D.}
Fig.~\ref{fig:nyu_quali} complements the cross-modal probing results with PCA visualizations of frozen features on an NYUv2 image.
For the 3D models, the image is lifted to 3D using MoGe-3~\citep{kong2026moge3}, a monocular geometry estimation model.
We additionally show DINOv3-B~\citep{simeoni2025dinov3} as a 2D reference.

\PARbegin{Unseen Indoor Scenes.}
Fig.~\ref{fig:replica_quali} visualizes VINO features on a large multi-room scene from Replica~\citep{straub2019replica}, which is not part of our pretraining data.
The PCA projection separates semantic regions consistently across the whole scene, and the cosine-similarity maps show that a query token activates other instances of the same category even when they are far apart.

\PARbegin{Outdoor Scenes.}
Fig.~\ref{fig:kitti_quali} shows VINO features on a LiDAR scan from SemanticKITTI~\citep{geiger2012kitti}.
A query on a pole activates other poles across the scan, and a query on a car activates other parked cars, suggesting that the representation captures object categories.

\end{document}